\documentclass[sigconf,authorversion]{acmart}
\usepackage{arydshln}
\usepackage{graphicx}
\usepackage{array} 

\usepackage{makecell}
\usepackage{diagbox}
\usepackage{graphicx}
\usepackage{amsmath}
\usepackage{booktabs}
\usepackage{blindtext}
\usepackage{caption}
\usepackage{float}
\usepackage{multirow}
\usepackage{url}
\usepackage{makecell}
\usepackage{multicol}

\usepackage{color}

\usepackage{bbding}
\usepackage{booktabs, tabularx}
\usepackage{xcolor}
\usepackage{xpatch}
\usepackage{newfloat}
\usepackage{listings}
\usepackage{array} 
\usepackage{colortbl}

\AtBeginDocument{%
  }

\renewcommand\footnotetextcopyrightpermission[1]{}
\usepackage{fancyhdr}
\setcopyright{acmlicensed}
\copyrightyear{2025}
\acmYear{2025}
\acmDOI{XXXXXXX.XXXXXXX}
\acmConference[MM"25]{ACM multimedia}{October 27--31,
  2025}{Dublin, Ireland}
\acmISBN{978-1-4503-XXXX-X/2018/06}

\begin{document}

\title{Category-Aware 3D Object Composition with Disentangled Texture and Shape Multi-view Diffusion}

\author{Zeren Xiong}
\affiliation{%
  \institution{Nanjing University of Science and Technology}
 \city{Nanjing}
  \country{China}}
\email{xzr3312@gmail.com}

\author{Zikun Chen}
\affiliation{%
  \institution{Nanjing University of Science and Technology}
 \city{Nanjing}
  \country{China}}
\email{zikunchencs@gmail.com}

\author{Zedong Zhang}
\affiliation{%
  \institution{Nanjing University of Science and Technology}
 \city{Nanjing}
  \country{China}}
\email{zandyz@njust.edu.cn}

\author{Xiang Li}
\affiliation{%
 \institution{Nankai University}
 \city{Tianjin}
 \country{China}}
\email{xiang.li.implus@nankai.edu.cn,}

\author{Ying Tai}
\affiliation{%
  \institution{Nanjing University}
  \city{Nanjing}
  \country{China}}
\email{yingtai@nju.edu.cn}

\author{Jian Yang}
\affiliation{%
  \institution{Nanjing University of Science and Technology}
 \city{Nanjing}
  \country{China}}
\email{csjyang@njust.edu.cn}

\author{Jun Li}
\authornote{Corresponding author.}
\affiliation{%
  \institution{Nanjing University of Science and Technology}
 \city{Nanjing}
  \country{China}}
\email{junli@njust.edu.cn}

\renewcommand{\shortauthors}{Xiong et al.}
\begin{abstract}
In this paper, we tackle a new task of 3D object synthesis, where a 3D model is composited with another object category to create a novel 3D model. However, most existing text/image/3D-to-3D methods struggle to effectively integrate multiple content sources, often resulting in inconsistent textures and inaccurate shapes. To overcome these challenges, we propose a straightforward yet powerful approach, category+3D-to-3D (C33D), for generating novel and structurally coherent 3D models. Our method begins by rendering multi-view images and normal maps from the input 3D model, then generating a novel 2D object using adaptive text-image harmony (ATIH) with the front-view image and a text description from another object category as inputs. To ensure texture consistency, we introduce texture multi-view diffusion, which refines the textures of the remaining multi-view RGB images based on the novel 2D object. For enhanced shape accuracy, we propose shape multi-view diffusion to improve the 2D shapes of both the multi-view RGB images and the normal maps, also conditioned on the novel 2D object. Finally, these outputs are used to reconstruct a complete and novel 3D model. Extensive experiments demonstrate the effectiveness of our method, yielding impressive 3D creations, such as \text{shark}(3D)-\text{crocodile}(text) in first row of Fig. \ref{fig:first}. 
\href{https://xzr52.github.io/C33D/}{\textcolor{magenta}{Project}}.
\end{abstract}


\begin{teaserfigure}
\centering
  \includegraphics[width=0.95\textwidth]{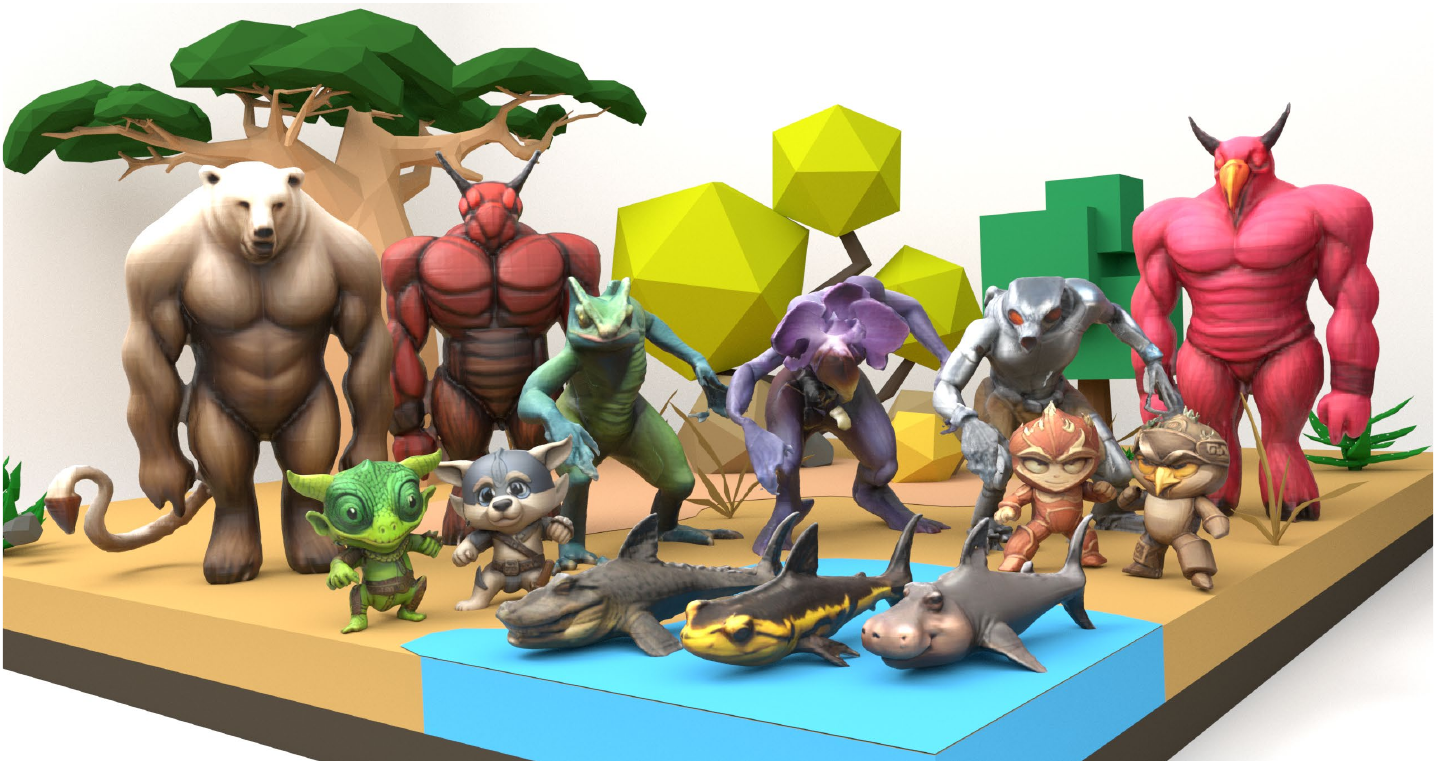}  \vskip -0in
  \caption{We propose a simple yet effective approach to create new 3D objects by combining a 3D model with another object category. Our algorithm successfully generates surprising and novel 3D models, such as \textit{devil} (3D)-\textit{ant} (text) in the last row.}
  \label{fig:first}
\end{teaserfigure}


\maketitle

\pagestyle{plain} 
\section{Introduction}
3D generation has attracted ongoing interest, particularly in the creation of novel 3D objects by merging multiple content sources, due to its broad applications in fields such as video games \cite{yang2023object,vanhoorick2023tracking,tendulkar2023flex}, animation \cite{abdal20233davatargan, animate3d,dhamo2025headgas,zhou2024headstudio}, and virtual reality \cite{zhou2024hugs,VR_3D} within computer vision and graphics. In contrast, this paper studies a 3D object synthesis task aimed at generating new 3D objects by blending an existing 3D model with a category of another object. 

Recent advances in text/image/3D-to-3D methods \cite{chen2024text,wang2024ThemeStation,chen2024Sculpt3D,liu2024Sherpa3D,li2024era3d} have demonstrated impressive capabilities in creating 3D objects from textual prompts or single 2D/3D inputs. These methods primarily aim to ensure consistency with input descriptions and visual fidelity while also enhancing diversity in the generated 3D samples. However, they remain limited in \textit{3D object synthesis}, particularly when integrating multiple content sources to produce novel combinations. For instance, ThemeStation \cite{wang2024ThemeStation} (3D-to-3D) generates thematically aligned 3D objects from exemplars but often fails to deliver a combinational surprise—a key aspect of novel design. Meanwhile, Era3D \cite{li2024era3d} (image-to-3D) employs multi-view diffusion to produce consistent images and normal maps in a canonical view, yet it sometimes yields incoherent or incomplete outputs (Fig. \ref{fig:mv_inversion}), leading to flawed 3D reconstructions. To overcome this issue, Sculpt3D \cite{chen2024Sculpt3D} and Sherpa3D \cite{liu2024Sherpa3D} (text-to-3D) incorporate coarse or sparse 3D shape priors. While these priors improve generation robustness, they often suffer from self-similarity and redundancy, hindering effective fusion of diverse textual and 3D inputs. 

To tackle this task, we develop \textbf{Category+3D-to-3D (C33D)} an effective method for 3D object synthesis that takes a 3D model and a target object category as inputs. The C33D pipeline consists of the following steps. Initially, the 3D model is rendered into multi-view RGB images (front, front-right, front-left, right, left, and back) along with their corresponding normal maps. Using adaptive text-image harmony (ATIH) \cite{Xiong2024ATIH}, we generate a novel 2D object from the front-view image and the target category. To ensure texture consistency across views, we introduce a \textit{texture multi-view diffusion} (TMDiff), which refines the remaining multi-view RGB images by conditioning them on the novel 2D object, injecting its key and value features into the self-attention module. Second, we propose a \textit{shape multi-view diffusion} (SMDiff), which refines the textured multi-view RGB images and normal maps by conditioning on the novel 2D object to input the diffusion, producing coherent multi-view outputs. We introduce a \textit{fusion-guided adaptive inversion} (FAI) method that dynamically adjusts the inversion step in SMDiff. By introducing a fusion score that is computed as the product of similarities between the synthesized 3D model, the input 3D model and the target category, FAI can ensure the synthesized model effectively preserves characteristics of both the input and the desired category. Finally, we use these outputs to render a complete novel 3D object. Our contributions are summarized as follows:
\begin{itemize}
\item To the best of our knowledge, we are the first to propose a category-aware 3D-to-3D synthesis method that blends a 3D model with a target object category, effectively generating novel 3D shapes with consistent texture and structure. 
\item We introduce a fusion-guided adaptive inversion method to dynamically adjust the inversion step in the diffusion process, preserving the features of both the input shape and the target category as effectively as possible.
\item Experimental results demonstrate the effectiveness of our method. Our approach shows superior performance in novel 3D object combination compared to the state-of-the-art image-to-3D and 3D-to-3D methods. Examples of these novel 3D objects, such as \textit{bear-greenhouse}, and \textit{rabbit-Aluminum} are shown in Figs. \ref{fig:first}, \ref{fig:experience1}, and \ref{fig:experience2}.
\end{itemize}

\begin{figure*}[t]
  \centering
 \includegraphics[width=0.91\linewidth]{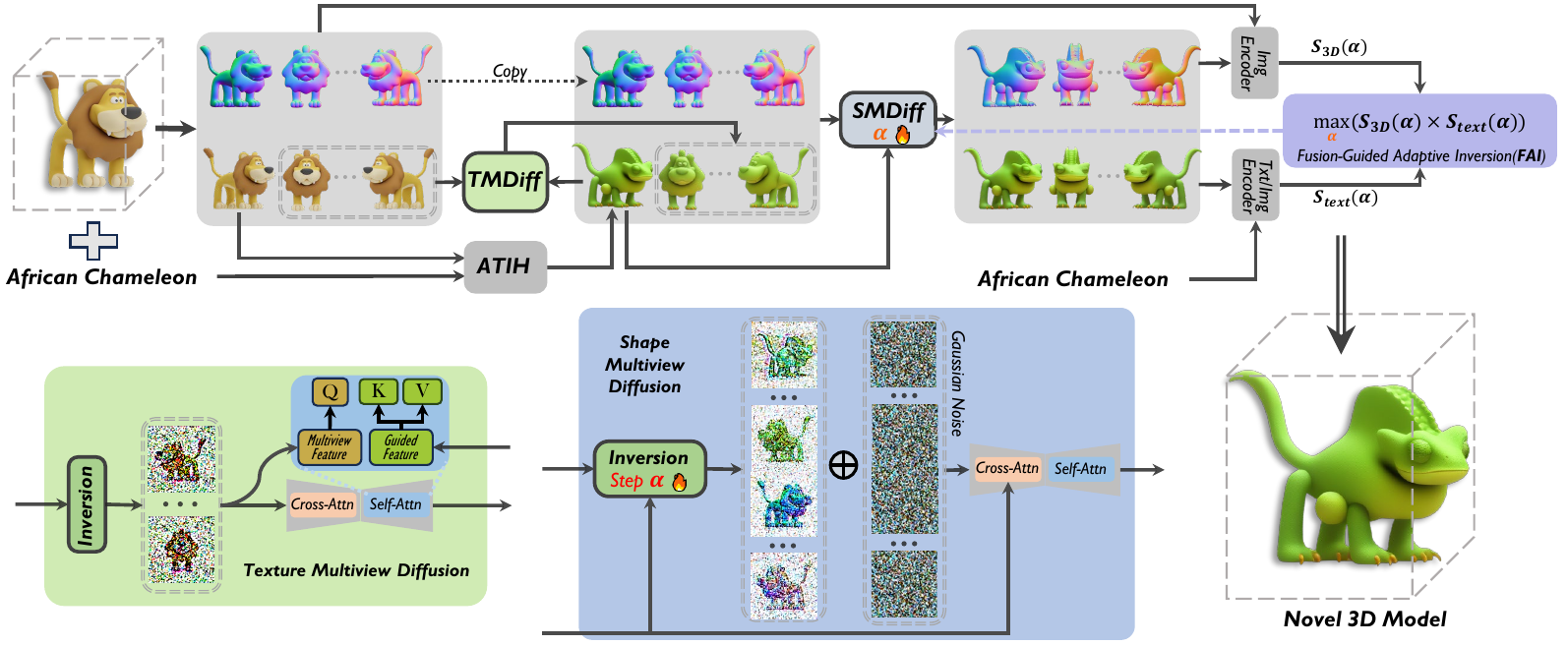}
 \vskip -0in
   \caption{Our category+3D-to-3D framework comprises  ATIH \cite{Xiong2024ATIH}, TMDiff, SMDiff, and FAI. ATIH generates a novel front-view 2D image by fusing the front-view image rendered from an input 3D model with an object category. TMDiff and SMDiff synthesize multi-view textures and shape/normal maps, respectively. FAI dynamically optimizes the inversion step ($\alpha$) via similarity maximization. The final output is a novel 3D model, rendered from inverted multi-view images and normal maps.}
   \label{fig:pipline}
\end{figure*}

\section{Related Work}
\label{sec:related_work}

\textbf{2D Semantic Mixing} focuses on generating a single object that integrates features from multiple concepts, and is key to creative image and object synthesis. Unlike multi-concept generation~\cite{kong2025omg,kwon2024concept,ding2024freecustom,multidiffusion} or style transfer~\cite{wang2023creativebirds,yang2022InST,chen2024artadapter,chung2024style,jiang2021deep}, it emphasizes concept-level fusion within one entity. 
Factorized Diffusion~\cite{geng2025factorized} separates frequency components of text prompts for semantic mixing. TP2O~\cite{li2024tp2ocreativetextpairtoobject} blends embeddings from multiple prompts to generate novel objects. ConceptLab~\cite{Richardson2024conceptlab} takes a lexical approach by interpolating between token embeddings. 
Image-based mixing methods such as MagicMix~\cite{liew2022magicmix} extract latents from input images and combine them with text during denoising, but often yield incoherent results. In contrast, ATIH~\cite{Xiong2024ATIH} achieves more harmonious image-text compositions using real image inputs. While most semantic mixing operates in 2D, our approach extends it to 3D, enabling fusion of real 3D models with textual semantics.

\textbf{Multiview Diffusion} aims to synthesize view-consistent images by leveraging cross-view information, enabling 3D-aware content generation. Recent works have explored this paradigm from multiple perspectives. Animate3D \cite{animate3d} and CAT3D \cite{cat3d} extend diffusion models to dynamic scene and object synthesis, while MVDiffHD \cite{MVDiffHD2024} enhances sparse-view reconstruction through high-resolution generation. LGM \cite{tang2024lgm} and CRM \cite{wang2024crm} introduce latent field and radiance field representations for higher-fidelity results, while Wonder3D \cite{long2024wonder3d} and MVFusion \cite{zhang2023mvf} aggregate multi-view signals to improve view consistency. In addition, Era3D \cite{li2024era3d} predicts camera parameters to mitigate distortion and produces high-resolution multi-view images. Despite these advances, existing methods still struggle to simultaneously preserve both the geometric structures and texture details across views. In contrast, our approach directly inverts multi-view renderings from a given 3D model and jointly models geometry and texture information, leading to more accurate and consistent multi-view synthesis.

\textbf{3D Object Editing} modifies an existing model based on natural language prompts, either locally or globally. Classical approaches rely on explicit geometry, including mesh deformations~\cite{yuan2021revisit,sorkine2005laplacian,sorkine2007rigid} and proxy-based editing~\cite{jacobson2012fast,yifan2020neural,sumner2005mesh}, which support fine-grained adjustments but lack flexibility for tasks like cross-category fusion or texture reconstruction.
Recent generative methods enable direct text-guided editing. Progressive3D~\cite{cheng2023progressive3d} restricts edits to spatial regions for better local alignment, while MVEdit~\cite{chen2024generic} edits multi-view images using 2D diffusion before 3D reconstruction. However, these methods often lack precision and fail to preserve semantic alignment, especially in text-object fusion scenarios.

\textbf{Exemplar-Based Generation} is widely used in 2D tasks such as image and scene synthesis~\cite{ruiz2023dreambooth,xia2025llmga,lu2025wovogen}, where outputs are conditioned on reference examples. This paradigm has been extended to 3D, allowing one-shot generation from a single image or model. TPA3D~\cite{wu2025tpa3d} and DreamBooth3D~\cite{raj2023dreambooth3d} enable fast or personalized 3D generation, while Patch3D~\cite{li2023patch} and Sin3DM~\cite{wu2023sin3dm} apply patch-based or diffusion methods to enhance realism. Phidias~\cite{wang2025phidias} improves diversity by fusing retrieved or user-provided references.  ThemeStation~\cite{wang2024ThemeStation} generates theme-aligned 3D content using two diffusion models per example, but remains constrained by the geometry and style of its exemplars. In contrast, our method fuses a 3D model and textual description to create structurally novel and semantically blended 3D content, enabling open-domain concept composition beyond exemplar-bound variation.

\section{Methodology}
Let \( M\) and \( T \) denote the input 3D object model and the target object category, respectively. Our objective is to generate a new 3D object model \( O\) by combining the 3D object model \( M \) with the object category \( T \). To accomplish this, we develop a Category+3D-to-3D (\textbf{C33D}) framework as illustrated in Figure~\ref{fig:pipline}. For convenience, \( M \) is rendered as multi-view images \( \{ I^s \}_{s\in \mathcal{S}}\) and normal maps \( \{ N^s \}_{s\in \mathcal{S}} \), 
where $\mathcal{S} = \{\textit{f, fr, fl, r, l, b}\}$\footnote{f: front, fr: front-right, fl: front-left, r: right, l: left, b: back}.
denotes a set of different views. 

\textbf{Overview.} We begin by using ATIH \cite{Xiong2024ATIH} to generate a novel front-view 2D object image \(I_{\text{nov}}^f \) by fusing the front-view image \( I^f\) and the object category \( T \).
In subsection~\ref{subsec:TMDiff}, then we introduce the \textit{Texture Multi-View Diffusion (TMDiff)} method, which transfers the texture features from \(I_{\text{nov}}^f \) to the remaining multi-view images \( \{ I^s \}_{s\in \mathcal{S}\backslash \{f\}}\), ensuring consistent textures across all viewpoints. Next, in subsection~\ref{subsec:SMDiff} we propose a \textit{Shape Multi-View Diffusion (SMDiff)}, which further integrates geometric and texture information from \(I_{\text{nov}}^f \) into the multi-view images provided by TMDiff and the normal maps for producing cohesive shape and texture synthesis. Additionally, subsection~\ref{subsec:evaluation_metric} introduces a \textit{fusion-guided adaptive inversion} to adaptively adjust the inversion step in SMDiff. Finally, we use the adjusted multi-view images and normal maps to render a novel 3D object model \( O\). 


\subsection{Texture Multi-View Diffusion}
\label{subsec:TMDiff}

Here, we present TMDiff to achieve consistent texture fusion between the front-view fused image \(I_{\text{nov}}^f \) and the multi-view images \( \{ I^{s} \}_{s\in \mathcal{S}\backslash \{f\}}\). Since we implement this Diffusion in latent space, we use an encoder \cite{sauer2024adversarial} to extract the latent features \(x_{\text{nov}}^f \) and  \( \{ x_0^s \}_{s\in \mathcal{S}\backslash \{f\}}\) from  \(I_{\text{nov}}^f \) and \( \{ I^s \}_{s\in \mathcal{S}\backslash \{f\}}\). 

TMDiff involves adding noise and performing denoising operations on the latent features. Given the current multi-view noisy features \( \{ x_{t}^s \}_{s\in \mathcal{S}\backslash \{f\}}\), the next-step multi-view features \( \{ x_{t-1}^s \}_{s\in \mathcal{S}\backslash \{f\}}\) are updated with \textbf{multi-view denoising operations}:
\begin{align}
x_{t-1}^s = &\nu_t x_t^s+ \beta_t \epsilon_\theta\left(x_t^s, t, \tau, \eta  \right) + \gamma_t \epsilon_t^s, \ \ \  s\in \mathcal{S}\backslash \{f\},
\label{eq:denoise}
\end{align}
where \( x_t^s\) represents the noise feature at step \(t\) for \(s\)-view, with \( \tau \) and \( \eta \) serving as conditioning inputs for cross-attention and self-attention, respectively, and \( \epsilon_t^s \) denoting standard Gaussian noise for the \(s\)-view. The coefficients \( \nu_t \), \( \beta_t \), and \( \gamma_t \) control the denoising weights, ensuring stability in the generation process. \( \epsilon_\theta(x_t^s, t, \tau, \eta) \) is the noise predicted by the pretrained U-net model \cite{sauer2024adversarial} under conditioning \(\tau=\text{Null}\) and \(\eta =x_{\text{nov}}^f \).

To transfer the texture of \(I_{\text{nov}}^f \) to each-view images \( I^s \), we draw inspiration from \cite{jeong2024visualstylepromptingswapping} and inject the feature \(x_{\text{nov}}^f \) into the key and value features within the UNet’s upsampling self-attention module for each view \( I^s \). This enables texture transfer during the denoising process without altering geometric structure. The multi-view self-attention under the condition \(\eta =x_{\text{nov}}^f\) is defined as:
\begin{align}
&\text{MSelfAttn}(x_t^s,\eta) = \text{Softmax}\left( \frac{Q_t^s (K^{\text{nov}})^\top}{\sqrt{d}} \right) V^{\text{nov}}, \ \ s\in \mathcal{S}\backslash \{f\},
\end{align} 
where \( d \) is the dimensionality of the key vectors, \( Q_t^s = x_t^s W_Q \), \( K^{\text{nov}} = x_{\text{nov}}^f W_K \), \( V^{\text{nov}} = x_{\text{nov}}^f W_V \), \( W_Q \), \( W_K \), and \( W_V \) are learned projection matrices.

Following the ReNoise inversion method \cite{garibi2024renoise}, under the denoising updation in Eq. \eqref{eq:denoise}, we have an approximation $\epsilon_\theta(x_{t}^s, t, \tau,\eta) \approx \epsilon_\theta(x_{t-1}^s, t, \tau,\eta) $ \cite{dhariwal2021diffusion}. Starting from the latent features \( \{ x_0^s \}_{s\in \mathcal{S}\backslash \{f\}}\), the \textbf{multi-view noise addition} step is reformulated as follows:
\begin{align}
x_{t}^s = (1/\nu_t)&\left(x_{t-1}^s - \beta_t \epsilon_\theta\left(x_{t-1}^s, t, \tau, \eta\right) - \gamma_t \epsilon_t^s\right), \ \ \ s\in \mathcal{S}\backslash \{f\},
\label{eq:addnoise}
\end{align} 
where \( \tau=\eta=\text{Null}\) represents a null condition. We further employ a decoder \cite{sauer2024adversarial} to reconstruct the multi-view images with consistent textures from the latent denoising features, denoted as \( \{ I_{\text{TMDiff}}^s\}_{s\in \mathcal{S}}\), where \(I_{\text{TMDiff}}^f=I_{\text{nov}}^f\). 

\begin{figure}[t]
  \centering
 \includegraphics[width=1\linewidth]{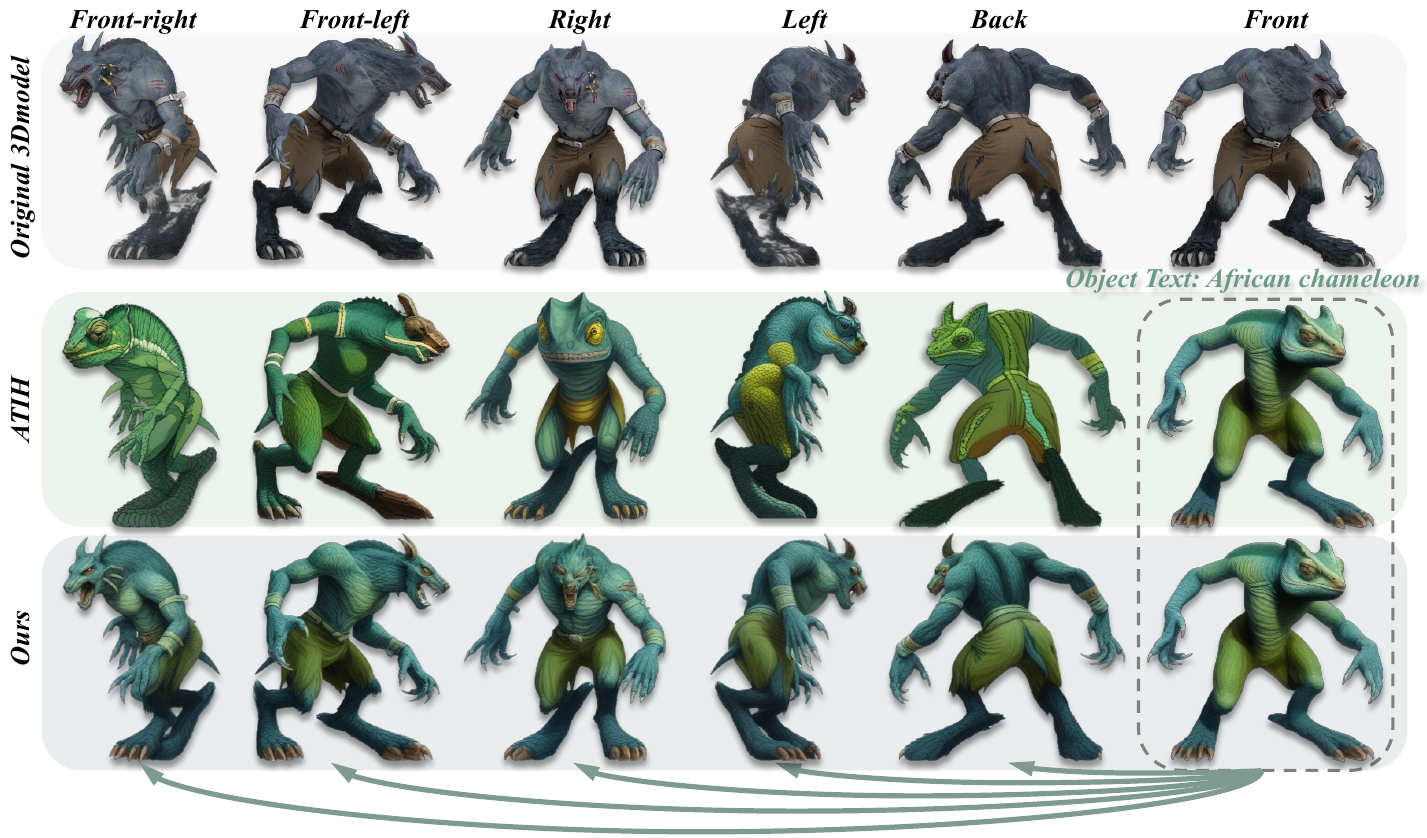}
 \vskip -0in
   \caption{An example of texture consistency. Our TMDiff achieves better texture consistency compared to ATIH \cite{Xiong2024ATIH}. }
   \label{fig:mvtexturealign}
\end{figure}

Our TMDiff method effectively utilizes the semantic richness of the self-attention layers to ensure texture consistency across views, leading to a higher level of visual coherence in the generated 3D object. For instance, as shown in the final row of Fig.~\ref{fig:mvtexturealign}, our method produces consistent textures across different views.

\textbf{Discussion 1.}
Ideally, ATIH \cite{Xiong2024ATIH} could be used to generate novel 2D multi-view images \( \{ I_{\text{nov}}^s \}_{s\in \mathcal{S}}\) by fusing \( \{ I^s \}_{s\in \mathcal{S}}\) and \( T \). However, as shown in the second row of Fig. \ref{fig:mvtexturealign}, this approach often leads to significant inconsistencies in texture and geometry across views, ultimately leading to failed 3D rendering. The underlying issue is that ATIH adaptively balances the fusion of each view image and the object text independently, leading to higher diversities in the generated \( \{ I_{\text{nov}}^s \}_{s\in \mathcal{S}}\).

\subsection{Shape Multi-View Diffusion}
\label{subsec:SMDiff}
After obtaining the multi-view images \( \{ I_{\text{TMDiff}}^s \}_{s\in \mathcal{S}}\), which exhibit consistent or highly similar textures to the novel front-view image
\(I_{\text{nov}}^f \), we observe that the shapes across these multi-view images (see Fig. \ref{fig:mvtexturealign}) and their corresponding normal maps \( \{ N^{s} \}_{s\in \mathcal{S}}\) remain inconsistent. To address this issue, we introduce SMDiff in this subsection, a method designed to achieve consistent shape fusion across the multi-view images \( \{ I_{\text{TMDiff}}^s \}_{s\in \mathcal{S}}\) and their associated normal maps \( \{ N^s \}_{s\in \mathcal{S}}\) using 
\(I_{\text{nov}}^f \) in latent space. Here, since \(I_{\text{TMDiff}}^f=I_{\text{nov}}^f\), we leverage this self-consistency to balance shape consistency across the multi-view images.

\begin{figure}[t]
  \centering
\includegraphics[width=0.87\linewidth]{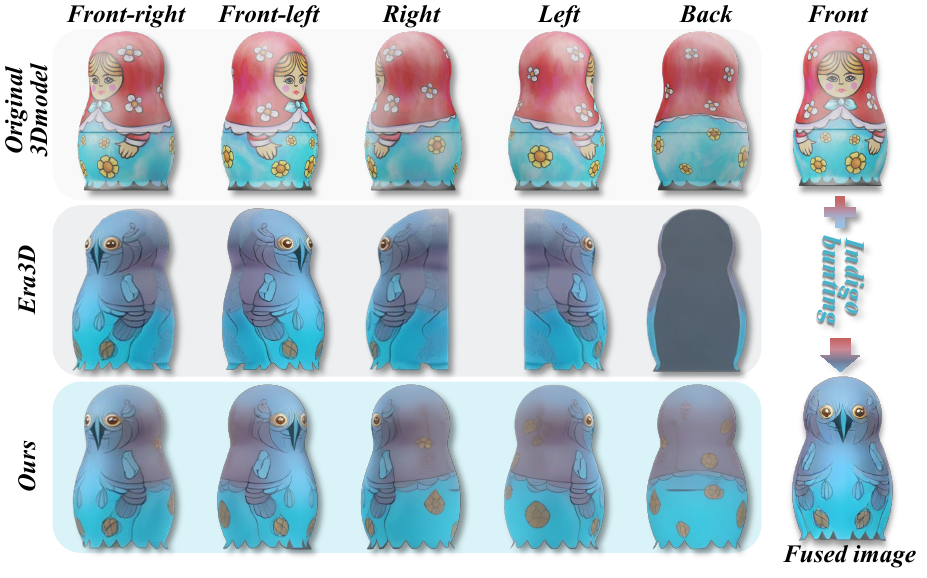}
 \vskip -0in
\caption{An example of shape accuracy. Our SMDiff demonstrates better shape accuracy compared to Era3D \cite{li2024era3d}. }
    \label{fig:mv_inversion}
\end{figure}

To achieve this, we first use an encoder \cite{li2024era3d} to extract the latent features \( \{ y_0^s, z_0^s \}_{s\in \mathcal{S}}\) from the multi-view images \( \{ I_{\text{TMDiff}}^s\}_{s\in \mathcal{S}}\) and normal maps  \( \{N^s \}_{s\in \mathcal{S}}\). The novel latent front-view feature \(x_{\text{nov}}^f \) has been extracted from \(I_{\text{nov}}^f \) in the last subsection. 
Similar to the TMDiff process, the SMDiff diffusion also involves a sequence of noise addition and denoising operations, conditioned on the feature 
\(x_{\text{nov}}^f \). Starting from the latent features \( \{ y_0^s, z_0^s\}_{s\in \mathcal{S}}\), the \textbf{multi-view noise addition} step follows Eq. \eqref{eq:addnoise} and is given by: 
\begin{equation}
\left\{\begin{aligned}
y_{t}^s = \left(y_{t-1}^s - \beta_t \epsilon_\theta\left(y_{t-1}^s, t, \tau_y, \eta\right) - \gamma_t \epsilon_{yt}^s\right)/\nu_t \\
z_{t}^s = \left(z_{t-1}^s - \beta_t \epsilon_\theta\left(z_{t-1}^s, t, \tau_z, \eta\right) - \gamma_t \epsilon_{zt}^s\right)/\nu_t
\end{aligned} 
\right., \ \ \  s\in \mathcal{S},
\label{eq:conditionaddnoise}
\end{equation}
where \( \epsilon_{yt}^s \) and \( \epsilon_{zt}^s \) represent the \(s\)-view standard Gaussian noise applied to the images and normal maps, respectively, \(\eta\) is a Null condition, and \(\tau_y\)/\(\tau_z\) are the prompt, \textit{a rendering image of 3D models, s-view, color/normal map}, to control the inversions of the multi-view images and normal maps. Upon reaching a specified inversion step \(t=\alpha\), we obtain the noised multi-view features \( \{ y_\alpha^s, z_\alpha^s \}_{s\in \mathcal{S}}\). \(t\) is controlled by the parameter \( \alpha \): when \( t < \alpha \), the inversion is used to progressively incorporate the geometric information of the 3D model \( M \) and the texture details from the novel front-view image \(I_{\text{nov}}^f\). Once \( t = \alpha \), the process transitions to the denoising diffusion, shifting the latent space toward a Gaussian distribution, resulting in smoother and more coherent multi-view images \cite{brooks2023instructpix2pix}.

To further enhance editability, we introduce additional Gaussian noise into the denoised latent features, as increasing the Gaussian noise level can improve shape editability \cite{Parmar2023zeroshotIIT}. Specifically, we modify the features as \( \{ y_\alpha^s+\epsilon_y^s, z_\alpha^s+\epsilon_z^s \}_{s\in \mathcal{S}}\), where \(\epsilon_y^s\) and \(\epsilon_z^s\) are the added Gaussian noise terms. Starting from these features, the next generation step involves updating the multi-view features \( \{ y_{t-1}^s, z_{t-1}^s \}_{s\in \mathcal{S}}\) using \textbf{multi-view denoising operations}: 
\begin{equation}
\left\{\begin{aligned}
y_{t-1}^s = &\nu_t y_t^s + \beta_t \epsilon_\theta\left(y_t^s, t, \tau_y, \eta \right)+ \gamma_t \epsilon_{yt}^s, \\
z_{t-1}^s = &\nu_t z_t^s + \beta_t \epsilon_\theta\left(z_t^s, t, \tau_z, \eta \right)+ \gamma_t \epsilon_{zt}^s, 
\end{aligned} 
\right. , \ \ \  s\in \mathcal{S},
\label{eq:conditiondenoise}
\end{equation}
where \(\eta=x_{\text{nov}}^f\) also indicates that \(x_{\text{nov}}^f\) serves as a condition to control the shape generation of the multi-view images and normal maps. The input features are refined by cascading the condition \(\eta\), expressed as \(y_t^s=y_t^s\copyright x_{\text{nov}}^f\) and \(z_t^s=z_t^s\copyright x_{\text{nov}}^f\). The input time embedding is enhanced by incorporating the embedding of the condition \(\eta\), defined as \(\text{Embedding}(t)=\text{Embedding}(t)+\text{Embedding}(x_{\text{nov}}^f) \).

\begin{figure}[t]
  \centering
\includegraphics[width=1.0\linewidth]{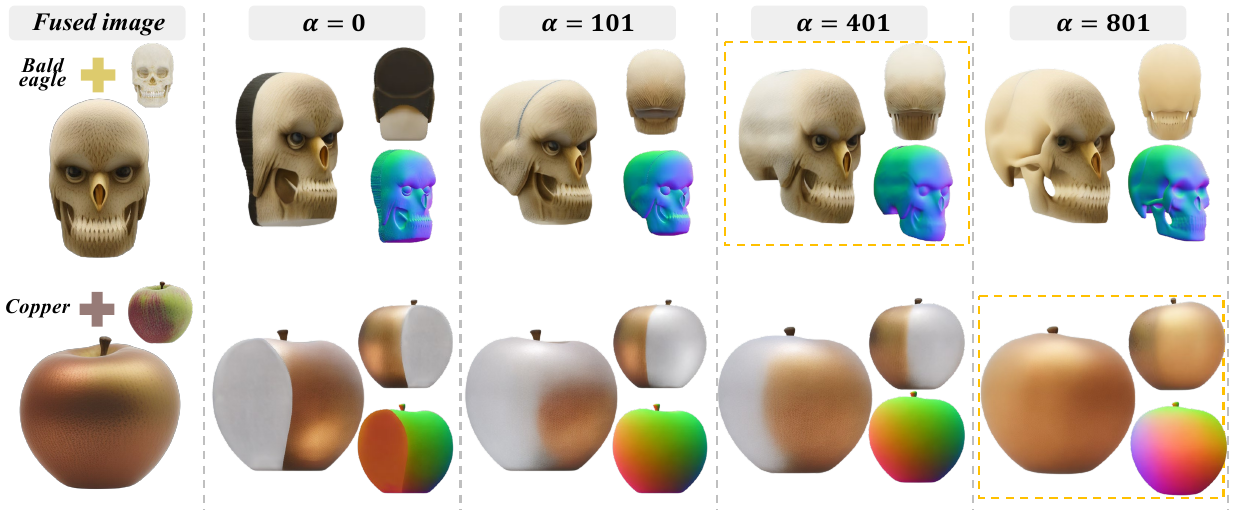}
 \vskip -0in
\caption{Different results with varying values of $\alpha$. Up and down 3D models are obtained by $\alpha=401$ and $801$, respectively. }
    \label{fig:Abalution_study_alpha}
\end{figure}

\textbf{Discussion 2.} Given the novel 2D fused object image \(I_{\text{nov}}^f \), a straightforward approach would be to apply an image-to-3D method to generate a new 3D object model. To demonstrate this, we use Era3D \cite{li2024era3d}, a state-of-the-art image-to-3D method. However, as illustrated in the second row of Fig.~\ref{fig:mv_inversion}, Era3D often produces geometric distortions and texture inconsistencies in the right, left, and back views. This results in unsuccessful 3D generation, as the model lacks sufficient geometric constraints from the input image.


\subsection{Fusion-Guided Adaptive Inversion}
\label{subsec:evaluation_metric}
In practice, we found that the fixed inversion step $\alpha$ in SMDiff often fails to  optimally balance preserving the visual characteristics of the 3D model \(M\) and integrating the target semantics from the object category \(T\). For example, Fig.~\ref{fig:Abalution_study_alpha} demonstrates that different 3D models typically require distinct inversion steps for optimal results. To address this limitation, we propose a Fusion-guided Adaptive Inversion (FAI) method, which dynamically adjusts the inversion step for better adaptation. Ideally, the output 3D model should neither simply replicate the original 3D model \(M\) nor completely sacrifice its structure to overfit the target semantics \(T\). Instead, it should achieve a balanced trade-off, preserving meaningful similarity to both \(M\) and \(T\). To quantify this trade-off, we define a \textbf{fusion score} at an inversion step \(\alpha\) as:
\begin{equation}
F(\alpha) = \sum_{s \in \mathcal{S}} w_s \cdot S_{3D}^s(\alpha) \cdot S_{text}^s(\alpha),
\label{eq:balancemetric}
\end{equation}
where \(S_{3D}^s(\alpha) \) measures the geometric and texture similarity between the output 3D model $O(\alpha)$ and the input 3D model \(M\) at $s$-view, while \(S_{text}^s(\alpha)\) evaluates the semantic similarity between the output 3D model $O(\alpha)$ and the target category \(T\) at $s$-view. We use their product to enforce that both terms must be sufficiently high to achieve a large fusion score. The weight \(w_s\) reflects the perceptual importance of each view (see Section~\ref{sec:experiments_settings} for details).

Specially, the 3D similarity is computed as:
\begin{align}
S_{3D}^{s}(\alpha) = & \lambda \cdot \cos(\phi(I^{s}_{\text{SMDiff}}(\alpha)), \phi(I^{s}))  \notag \\
& + (1-\lambda) \cdot \cos(\phi(N^{s}_{\text{SMDiff}}(\alpha)), \phi(N^{s}))
\end{align}
where $\phi(\cdot)$ denotes the DINO-extracted feature \cite{oquab2024dinov2}, $cos(\cdot)$ represents the cosine similarity between two feature vectors, and \(\lambda\) (set to 0.5) control their relative importance. The $s$-view image \(I^{s}_{\text{SMDiff}}(\alpha) \) and normal maps \( N^{s}_{\text{SMDiff}}(\alpha) \) are generated by using SMDiff, while the corresponding $s$-view image \( I^s  \) and normal maps \( N^s \) are rendered from the input 3D model $M$.
Similarly, the semantic similarity is computed as:
\begin{align}
S_{text}^s(\alpha)  = cos(\text{CLIP}(I_{\text{SMDiff}}^s(\alpha)),\text{CLIP}(T)),
\end{align}
using CLIP~\cite{radford2021learning} features based on cosine similarity.

\textbf{Adaptive Inversion.} To determine the optimal inversion step \(\alpha\), we perform ternary search \cite{arora2016novel} over \([1, 901]\) in increments of 100 steps, i.e., \(\alpha \in \{1, 101, 201, \dots, 901\}\). This balances efficiency with the observed fusion dynamics: larger $\alpha$ preserves more geometric details from the input 3D model $M$, while smaller $\alpha$
emphasizes semantic guidance from the category prompt $T$. Formally, the optimal \(\alpha^*\) is selected by maximizing the fusion score defined in Eq.~\eqref{eq:balancemetric}:
\begin{equation}
\alpha^\ast = \arg\max_\alpha {F}(\alpha).
\end{equation}

This adaptive strategy allows our framework to automatically balance structural preservation and semantic fusion, effectively guiding SMDiff to generate multi-view results that maintain both geometric consistency and semantic relevance for reconstruction.

\begin{figure*}[t]
  \centering
 \includegraphics[width=0.84\linewidth]{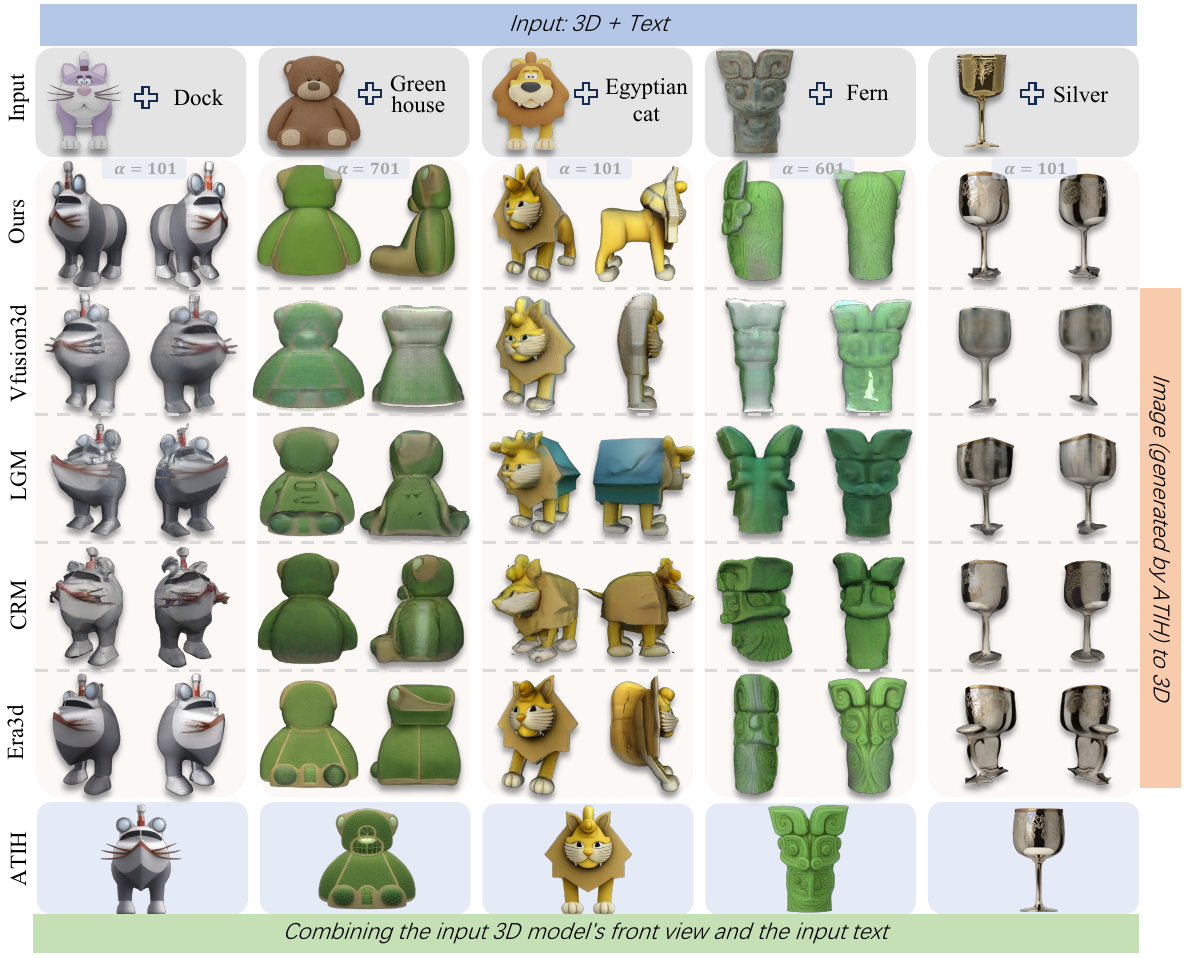}
   \caption{Comparison with existing image-to-3D methods—using images generated by ATIH \cite{Xiong2024ATIH}. We observe that Era3D \cite{li2024era3d}, CRM \cite{wang2024crm} and LGM \cite{tang2024lgm}, and Vfusion3D \cite{han2024VFusion3D} struggle with inconsistent textures and inaccurate shapes in the generated 3D object models. In contrast, our method successfully synthesizes novel 3D objects, such as the \textit{Cat-Dock} and \textit{Tiger-Egyptian Cat}, shown in the first and fourth columns, respectively.}
   \label{fig:experience1}
\end{figure*}

\textbf{Fusion-based evaluation metric.} To further assess the overall fusion quality of the output 3D object \(O\)—rendered from the generated multi-view images and normal maps—we propose a simple yet effective metric inspired by Eq.~\eqref{eq:balancemetric}. Specifically, the evaluation score is computed as:
\begin{equation}
F_{\text{sim}} = S_{3D}(O,M) \times S_{text}(O,T),
\label{eq:eval_metric}
\end{equation}
where the semantic alignment score \(S_{text}(O,T) = \text{CLIP}(O, T)\) quantifies the similarity between \(O\) and \(T\), computed as the average over multi-view renderings. The geometric similarity \(S_{\text{3D}}\) evaluates shape and texture consistency, defined as follows:
\begin{equation}
S_{3D}(O,M) = \text{mean}(S_{\text{geo}}, S_{\text{texture}}),
\end{equation}
where the geometric similarity \(S_{\text{geo}}\) is measured by \(F_{\text{score}}\)~\cite{Sokolova2006BeyondAF} between \(O\) and \(M\), while the texture similarity \(S_{\text{texture}}\) is computed as the average DINO~\cite{oquab2024dinov2} similarity across their multi-view renderings. Both \(S_{3D}(O,M)\) and \(S_{text}(O,T)\) are normalized to \((0,1)\), ensuring that \(F_{\text{sim}}\) effectively captures the overall alignment quality between \(M\) and \(T\). A higher $F_{\text{sim}}$ score indicates a better output $O$.

\section{Experiments}
\label{sec:experiments}

\begin{figure*}[t]
  \centering
 \includegraphics[width=0.83\linewidth]{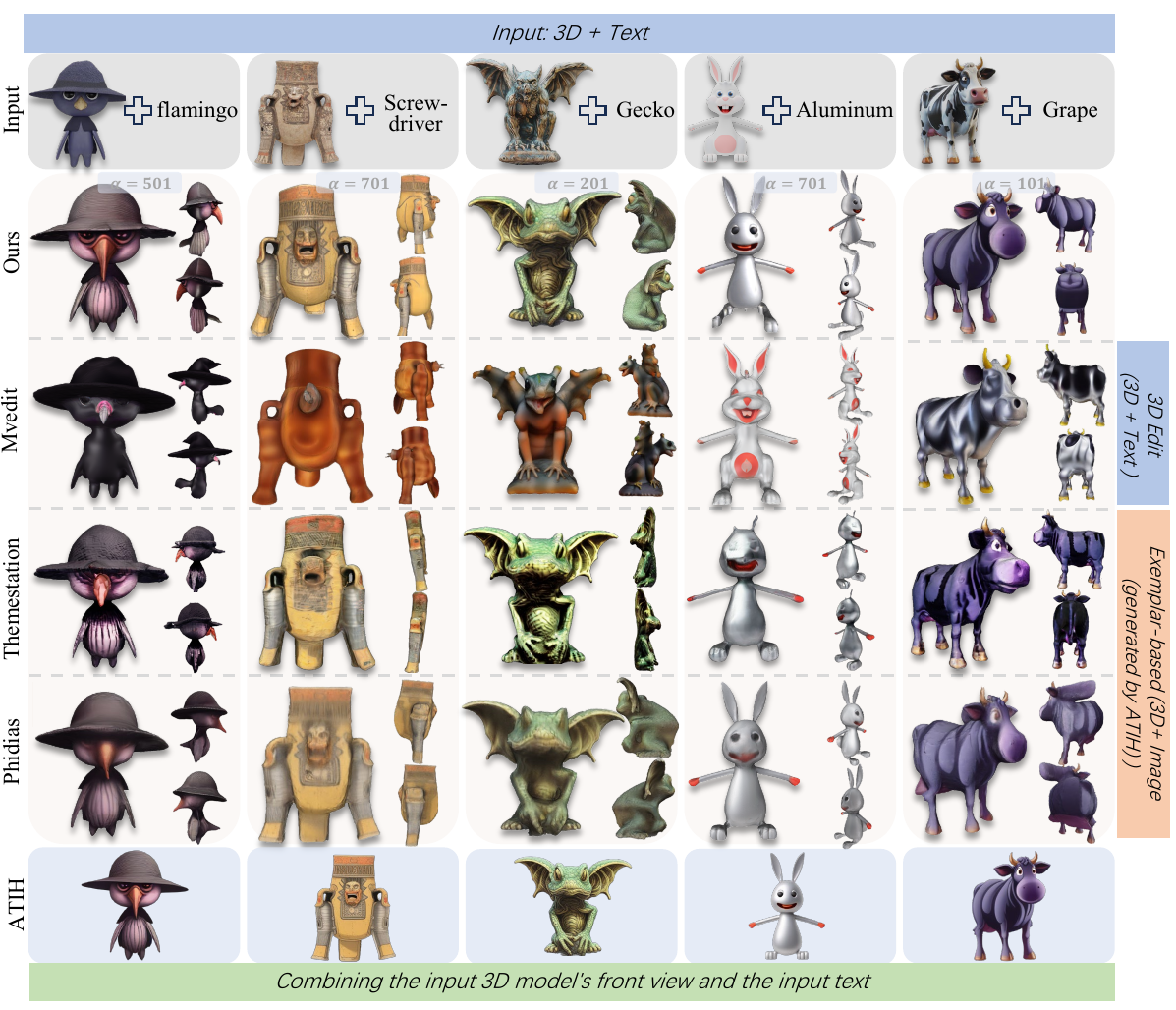}
   \caption{\textbf{Comparisons with 3D-to-3D methods.}  We observe that ThemeStation \cite{wang2024ThemeStation} and Phidias\cite{wang2025phidias} often produce incomplete shapes and blurry textures, while MvEdit \cite{chen2024generic} tends to over-preserve the original shape with limited semantic fusion. In contrast, our C33D achieves more accurate shapes, richer textures, and better cross-view consistency.}
   \label{fig:experience2}
\end{figure*}

\subsection{Experimental Settings}
\label{sec:experiments_settings}
\textbf{Datasets}. We constructed a dataset of 110 pairs of 3D models and text descriptions for model evaluation and testing. These 3D models, sourced from Microsoft 3D Viewer\footnote{\url{https://en.wikipedia.org/wiki/Microsoft_3D_Viewer/}} and Sketchfab\footnote{\url{https://sketchfab.com/}}, include 22 models in three categories: animals, objects, and cartoon characters. Each model has 100 associated descriptions across five main object categories: architecture, tools, metal, plants, and animals (see Appendix \ref{sec:datasets}). To enhance diversity in object descriptions, we used ChatGPT \cite{openai2023chatgpt} to generate 20 unique descriptions per category, covering a variety of scenes and features. Using the ATIH fusion method, we combined the primary view of each 3D model with its corresponding description to create 2D images.
To ensure dataset quality, we evaluated images with aesthetic scores (AES) \cite{aestheticcode} and human preference scores (HPS) \cite{wu2023hpsv2}. Based on these scores, we selected the most representative images for the final test dataset.

\textbf{Metrics}. 
To thoroughly evaluate our method, we selected six metrics to assess the quality of generated images in terms of visual quality, semantic alignment, and geometric consistency. These metrics include Aesthetic Score (AES) \cite{aestheticcode}, CLIP Text-Image Similarity (CLIP-T) \cite{radford2021learning}, DinoV2 Image Similarity (Dino-I) \cite{oquab2024dinov2}, Human Preference Score (HPS) \cite{wu2023hpsv2}, Geometric Similarity \(F_{\text{score}}\) \cite{ghosh2015fscore} (thresholded at 0.02), our fusion evaluation metric \(F_{\text{sim}}\), which is formulated in Eq.~\ref{eq:eval_metric}. These metrics together provide a comprehensive evaluation of the model quality.

\textbf{Implementation details}. Our method was implemented on both SDXLturbo \cite{sauer2024adversarial} and Era3D \cite{li2024era3d}, with all input images resized to \(512 \times 512\)  pixels for consistency across experiments and fair comparison. The inversion process used the Ancestral-Euler sampler\cite{karras2022elucidating} for SDXLturbo and the DDIM sampler \cite{ddim} for Era3D. During the Fusion-Guided Adaptive Inversion process, to balance multi-view consistency and prevent overfitting to the front view, we set the weight of the front view to \(w_f=0.10\), while assigning \(w_{fr}=w_r=w_b=w_l=w_{fl}=0.18\) to the other views. For multi-view 3D model reconstruction, we followed Wonder3D \cite{long2024wonder3d}, using NeuS for initial reconstruction and then applying texture refinement to improve visual fidelity. All experiments were conducted on two NVIDIA GeForce RTX 4090 GPUs.

\subsection{Main Results}
To evaluate the effectiveness of our proposed C33D framework, we conduct systematic comparisons with several state-of-the-art 3D generation methods. Specifically, we select four image-to-3D approaches, including Era3D~\cite{li2024era3d}, LGM~\cite{tang2024lgm}, CRM~\cite{wang2024crm}, and Vfusion3D~\cite{han2024VFusion3D}, and three recent 3D-to-3D generation methods: ThemeStation~\cite{wang2024ThemeStation}, Phidias~\cite{wang2025phidias}, and Mvedit~\cite{chen2024generic}. These baselines cover both image-to-3D and 3D-to-3D tasks, representing the current leading techniques in 3D content generation.


\begin{table}[htbp]
\vskip -0in
\centering
\setlength{\tabcolsep}{7pt}
\renewcommand{\arraystretch}{1.1}
\caption{Quantitative comparisons on our dataset.}
\vskip -0in
\label{tab:quant_cmp}
\vskip -0in
\resizebox{0.95\linewidth}{!}{
\begin{tabular}{c||c|c|c|c|c}
\Xhline{1.2pt}
Models & DINO-I$\uparrow$ \cite{oquab2024dinov2}     & CLIP-T$\uparrow$ \cite{radford2021learning}      & AES $\uparrow$ \cite{aestheticcode}   & HPS$\uparrow$ \cite{wu2023hpsv2}     & $F_{\text{sim}}$$\uparrow$ \\ \hline
\rowcolor{green!10} \textbf{Ours}           & \textcolor{red}{\textbf{0.715}}  & 0.299  & \textcolor{red}{\textbf{ 4.475}} & \textcolor{blue}{0.347}       & \textcolor{red}{\textbf{0.312}}\\
CRM \cite{wang2024crm}        & 0.683  & 0.302  & 4.290 & 0.331       & 0.247\\
Era3D \cite{li2024era3d}         &  0.675 & 0.298 & \textcolor{blue}{4.466}    & 0.343      & 0.217 \\
LGM \cite{tang2024lgm}        & 0.665   & \textcolor{red}{\textbf{0.304}} & 4.405 & 0.336       & 0.245\\
Vfusion3d \cite{han2024VFusion3D} & 0.490 & 0.293   & 4.162& 0.318      &0.192 \\
Themestation \cite{wang2024ThemeStation} & \textcolor{blue}{0.698} &  0.302   & 4.350& \textcolor{red}{\textbf{0.350}}   &0.294\\
Phidias \cite{wang2025phidias} & 0.685 &  \textcolor{blue}{0.303}   & 4.358 &0.337    &0.261\\
Mvedit \cite{chen2024generic} & 0.683 &  0.286   & 4.396 &0.327     & \textcolor{blue}{0.309}\\
\Xhline{1.2pt}
\end{tabular}}
\vskip -0in
\end{table}


\begin{table*}[t]
\centering
\setlength{\tabcolsep}{7pt}
\renewcommand{\arraystretch}{1.1}
\caption{User study with image-to-3D and 3D-to-3D methods.}
\vskip -0in
\label{tab:user_stu}
\resizebox{0.87\linewidth}{!}{
\begin{tabular}{c||c|c|c|c|c||c|c|c|c}
\Xhline{1.2pt}
       & \multicolumn{5}{c||}{image-to-3D}      & \multicolumn{4}{c}{3D-to-3D}  \\
       \hline
Models & Ours & CRM \cite{wang2024crm}  & Era3D \cite{li2024era3d} & LGM \cite{tang2024lgm} &  Vfusion3D \cite{han2024VFusion3D} & Ours & Themestation \cite{wang2024ThemeStation}   & Phidias \cite{wang2025phidias}  & Mvedit \cite{chen2024generic}             \\
\hline
Vote $\uparrow$   & \textbf{355}  & 36        & 66 &  24   & 9  & \textbf{361}  & 33      & 80 & 16                     \\
\Xhline{1.2pt}
\end{tabular}}
\end{table*}

\textbf{Comparison with Image-to-3D Methods.}  
For image-to-3D methods, we use the fused image \(I_{nov}^f\) as input, and the qualitative comparisons are shown in Fig.~\ref{fig:experience1}. Existing methods, such as Vfusion3D, LGM, and Era3D, often suffer from the \textit{Janus problem}, where the generated 3D shapes exhibit overly similar appearances between the front and back views. This issue leads to unnatural results, as the side views of the object may appear like two mirrored front views stitched together (e.g., the fusion of ``teddy bear'' and ``greenhouse'').
CRM mitigates this issue by improving shape accuracy but still struggles to preserve fine-grained details, often resulting in overly smooth surfaces and blurry textures, as shown in the third column. Although Era3D achieves better texture fidelity, its limited shape control often introduces geometric distortions or unnatural shape protrusions, as observed in the fifth column.
In contrast, our C33D method explicitly leverages the structure and appearance of the original 3D model, enabling accurate shape reconstruction, rich detail preservation, and view-consistent texture generation across multiple views.

\textbf{Comparisons with 3D-to-3D Methods.}
We compare our C33D framework with two representative exemplar-based 3D generation methods, ThemeStation and Phidias, as well as a 3D editing method, MvEdit, as shown in Fig.~\ref{fig:experience2}.
For ThemeStation and Phidias, we follow their pipelines by using the original 3D model together with the images generated by ATIH as input for reconstruction. As shown in Fig.~\ref{fig:experience2}, both methods suffer from varying degrees of shape incompleteness and texture blurring, especially in the examples of the third and fifth columns.For the 3D editing method MvEdit, we follow its input format by providing the original 3D model along with a text prompt in the form of ``creative fused with a <object category>''. As shown in Fig.~\ref{fig:experience2}, MvEdit often fails to effectively integrate the original 3D model with the target concept, producing results that remain overly similar to the original shape with limited semantic modification. In comparison, our C33D method achieves a more harmonious fusion of geometry and semantics, resulting in 3D objects with consistent textures and coherent structural changes across multiple views.

\textbf{Quantitative Results.}
As shown in Tab.~\ref{tab:quant_cmp}, our C33D consistently outperforms all baselines across AES, $F_{\text{sim}}$, and DINO-I scores. This demonstrates its ability to achieve high-fidelity fusion between the input 3D model and target category while preserving visual quality, geometric integrity, and fine structural details. Among baselines, LGM achieves competitive CLIP-T scores, indicating strong semantic alignment with the category text. However, its lower DINO-I and $F_{\text{sim}}$ scores reveal limitations in geometry preservation and harmonious fusion. MvEdit ranks the second-best $F_{\text{sim}}$ score, suggesting better shape retention, but its poor CLIP-T score reflects weak semantic fusion—often producing results too similar to the input with minimal text-driven changes. Compared to ThemeStation, C33D excels in both overall quality and fusion effectiveness. While ThemeStation achieves marginally higher HPS scores, its outputs exhibit geometric inconsistencies and lack fine details, particularly when contrasted with our method.

\subsection{User Study}

To evaluate the perceptual quality of our results, we conducted two user studies under different comparison settings. As summarized in Table~\ref{tab:user_stu}, the left part compares our method with existing image-to-3D approaches, while the right part compares our C33D method against several 3D-to-3D methods sharing a similar problem setup. Each study includes five question groups, with visual examples illustrated in Figure~\ref{fig:experience1} and Figure~\ref{fig:experience2}. In total, we collected 490 responses from 98 participants for each study. In both comparisons, our C33D consistently received the highest user preference. 

Specifically, in the image-to-3D comparison, C33D obtained 72.45\% of the total votes, substantially surpassing LGM~\cite{tang2024lgm} (4.90\%), Era3D~\cite{li2024era3d} (13.47\%), Vfusion3D~\cite{han2024VFusion3D} (1.84\%), and CRM~\cite{wang2024crm} (7.35\%). In the 3D-to-3D comparison, our method achieved a dominant 73.67\% vote share, significantly outperforming Themestation~\cite{wang2024ThemeStation} (6.73\%), Phidias~\cite{wang2025phidias} (16.33\%), and Mvedit~\cite{chen2024generic} (3.27\%). These results clearly demonstrate that users strongly prefer C33D over both image-to-3D baselines and 3D-to-3D approaches, highlighting the superior perceptual quality and effectiveness of our framework.

\subsection{ Ablation Study}

\begin{figure}[t]
  \centering
\includegraphics[width=0.97\linewidth]{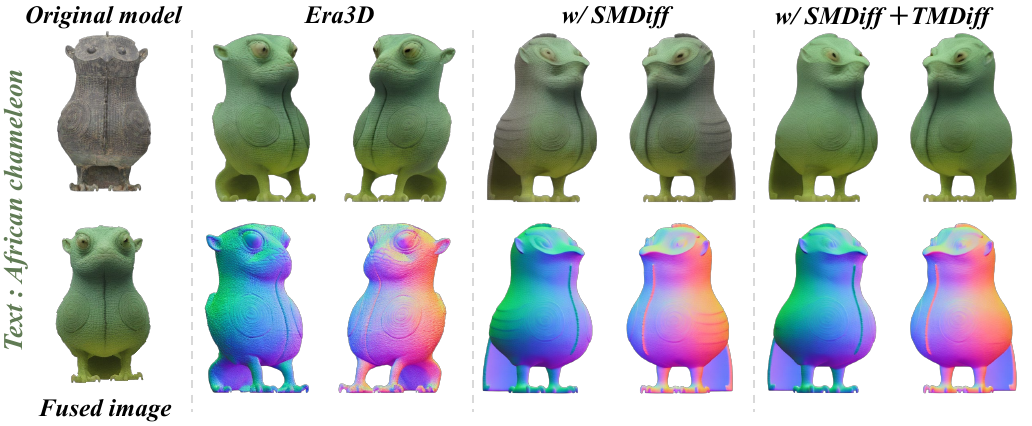}
 \vskip -0in
\caption{Ablation study of SMDiff and TMDiff.}
    \label{fig:Abalution_study_2}
\end{figure}
Fig.~\ref{fig:Abalution_study_2} presents the ablation results of our two key modules, SMDiff and TMDiff. The leftmost column shows the original 3D model and the target fused image. As shown, the results of Era3D suffer from obvious shape incompleteness due to the lack of structural priors. By introducing the SMDiff module, this issue can be alleviated by leveraging the geometric information from the original model. However, when there is a significant texture mismatch between the model and the fused image, noticeable texture inconsistency still occurs, such as unharmonious colors that are inconsistent with the target image. By further introducing the TMDiff module and adaptively searching for the optimal $\alpha$, our method can significantly mitigate these issues, producing more consistent textures and visually harmonious results across different views.

\section{Conclusion}
\label{sec:conclusion}

In this paper, we proposed a novel 3D object synthesis framework that fuses 3D models with textual descriptions to create unique and surprising 3D objects. We presented texture and shape multi-view diffusion to generate novel multi-view images and corresponding normal maps, which are then used to render the new 3D model. Extensive experiments demonstrated the effectiveness of our framework in producing a diverse range of visually striking 3D object fusions. This capability is especially valuable for designing innovative and captivating 3D animated characters in the entertainment and film industries. \textit{The limitation is provided in Appendix \ref{sec:limitation}.}

\bibliographystyle{ACM-Reference-Format}
\bibliography{main_acmm}

\clearpage

\appendix

The supplementary materials provide a detailed analysis of the experimental setup, results, and user studies, further validating the effectiveness and innovation of our method. \textbf{Section~\ref{sec:datasets}} outlines the dataset selection criteria and describes the text and 3D model classifications used, ensuring evaluation diversity and representativeness. \textbf{Section~\ref{sec:Implementation}} discusses the experimental setup, including model configurations and sampling techniques. In \textbf{Section~\ref{sec:limitation}}, we acknowledge the limitations of our method, particularly the redundant mesh surfaces from neural SDF reconstruction, and propose integrating MeshAnything \cite{chen2024meshanythingv2artistcreatedmesh} to improve mesh generation and performance. \textbf{Section~\ref{sec:User study}} presents user study results, confirming the superiority of our method in 3D model fusion and artistic expressiveness. Additionally, we provide More Ablation Studies in \textbf{Section~\ref{sec:More Ablaution}} to thoroughly evaluate the impact of core components in our pipeline, and More Comparisons in \textbf{Section~\ref{sec:More Comparisons}} to demonstrate the robustness of our approach against other methods. Finally, \textbf{Section~\ref{sec:More results}} demonstrates the versatility and robustness of our method in generating creative 3D models.

\section{Datasets}
\label{sec:datasets}

\begin{figure}[htbp]
  \centering
\includegraphics[width=0.90\linewidth]{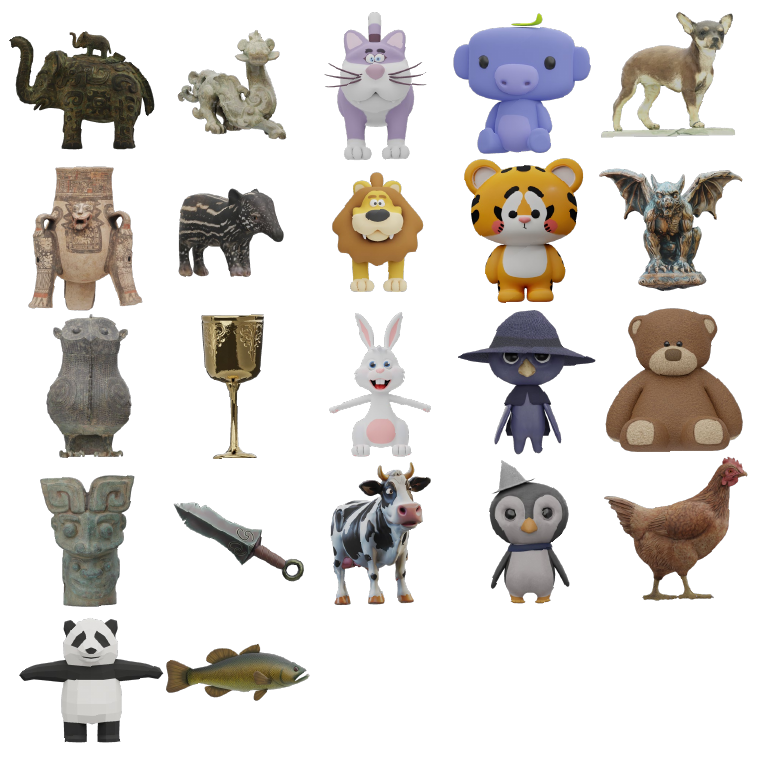}
\caption{Original 3D model Set. }
    \label{fig:sup_img_set}
\end{figure}

We selected 100 text descriptions (as detailed in Table \ref{tab:sup_text_cat}) and divided them into five groups, each containing texts with similar themes to ensure both representativeness and diversity in classification. As shown in Figure \ref{fig:sup_img_set}, the 22 models correspond to different text categories (further detailed in Table \ref{tab:sup_model_cat}). Experimental results demonstrate that our method effectively achieves content fusion between 3D models and text across different categories, showcasing exceptional generalization capability.

\begin{table}[htbp]
\vskip -0in
\centering
\setlength{\tabcolsep}{7pt}
\caption{Original 3D model Categories.}
\renewcommand{\arraystretch}{1.1}
\scalebox{1.0}{
    \begin{tabular}{>{\raggedright}m{1.5cm}||>{\raggedright\arraybackslash}m{5cm}}
        \Xhline{1.2pt}
        \textbf{Category} & \textbf{Items} \\
        \hline
        Objects & gargoyle statue, goblet, jaguar vessel, lidded ewer, pole finial, sword, tapir baby, bear, winged dragon, owl zun  \\
        \hline
        Animals & chicken, dog, fish \\
        \hline
        Cartoon characters & cartoon cannon, cartoon cow, cartoon crow, mascot, cartoon lion, cartoon panda, cartoon cat, cartoon penguin, cartoon tiger \\
        \Xhline{1.2pt}
    \end{tabular}
    }

\label{tab:sup_model_cat}
\end{table}

\begin{table*}[htbp]
\vskip -0in
\centering
\setlength{\tabcolsep}{7pt}
\renewcommand{\arraystretch}{1.1}
\caption{List of Text Items by Object Category.}
\vskip -0.1in
\label{tab:sup_text_cat}
\resizebox{0.82\linewidth}{!}{
\begin{tabular}{>{\raggedright}m{3cm}||>{\raggedright\arraybackslash}m{10cm}}
        \Xhline{1.2pt}
        \textbf{Category} & \textbf{Items} \\
        \hline
        Animals & zebra, peacock, giraffe, polar bear, flamingo, triceratops, penguin, Egyptian cat, tiger, elephant, sea turtle, shark, butterfly, orangutan, spider, whale, owl, snail, gecko, ant \\
        \hline
        Plants & cactus, sunflower, oak tree, rose, bamboo, sycamore tree, lotus, fern, pineapple, tea tree, palm tree, strawberry, tomato, pumpkin, grape, orchid, onion, lily, pine tree, banana tree \\
        \hline
        Metals & brass, gold, steel, aluminum, copper, silver, platinum, titanium, nickel, zinc, lead, iron, stainless steel, magnesium, tungsten, tin, molybdenum, chromium, cobalt, palladium \\
        \hline
        Tools & hammer, screwdriver, electric drill, pliers, wrench, scissors, measuring tape, level, welder, cutter, chainsaw, peeler, espresso maker, blender, oven, mop, flashlight, fan, sandpaper, ruler \\
        \hline
        Buildings & lighthouse, barn, greenhouse, clock tower, theater, subway station, stadium, gas station, bridge, tunnel, dock, pavilion, ferris wheel, monument, temple, tower, transmission tower, windmill, amphitheater, watermill \\
        \Xhline{1.2pt}
\end{tabular}}
\end{table*}

\section{Limitation}
\label{sec:limitation}
Although C33D successfully achieves the fusion of 3D models and text to generate novel 3D models, the reconstruction algorithm used in our current method is based on neural SDF, resulting in a certain level of redundancy in the number of mesh surfaces extracted. This redundancy makes the generated models less efficient in terms of detail representation and computational resource consumption, limiting their direct applicability to practical industrial scenarios. 

Thus, while our method has achieved notable results in academic research, further optimization is necessary to meet the requirements of industrial-grade applications. To address this, we plan to integrate our framework with MeshAnything \cite{chen2024meshanythingv2artistcreatedmesh} in future work, enabling the direct generation of more streamlined and efficient industrial-grade meshes. This integration will enhance both the usability and performance of the generated models in real-world applications.

\section{Implementation Details}
\label{sec:Implementation}
In our C33D framework, we render both RGB images and normal maps from six predefined viewpoints \([0^\circ, 45^\circ, -45^\circ, -90^\circ, 90^\circ, 180^\circ]\) of the input 3D model to capture diverse geometric and appearance information.

For ATIH \cite{Xiong2024ATIH}, we follow the default configuration using SDXLturbo \cite{sauer2024adversarial} as the base model. The fusion image \(I_{\text{nov}}^f\) is generated by employing the golden-section search strategy to automatically find the optimal text-image harmony.
In our proposed TMDiff, we adopt the Ancestral-Euler sampler \cite{karras2022elucidating} for inverting the multi-view RGB images. During the denoising process, we inject the feature of \(x_{\text{nov}}^f\) into layers \([128, 140]\) of the Unet at steps \([3, 4]\), ensuring texture consistency across different views.
For our proposed SMDiff, we build upon Era3D \cite{li2024era3d} as the backbone model and apply the DDIM sampler \cite{ddim} to jointly invert all multi-view images. The inversion process is conducted within the timestep range of \([0, 1000]\), while the inversion step length \(\alpha\) is adaptively tuned to precisely control the degree of information retention from the input 3D model.
During the Fusion-Guided Adaptive Inversion process, we assign the view weights to balance multi-view consistency and prevent overfitting to the front view. Specifically, the front view weight is set to \(w_f=0.10\), while the remaining five views share equal weights \(w_{fr}=w_r=w_b=w_l=w_{fl}=0.18\). Empirically, the ternary search strategy typically converges to the optimal \(\alpha\) within five iterations, demonstrating its efficiency and stability in balancing the trade-off between shape preservation and semantic fusion.
\section{User study}
\label{sec:User study}
In this section, we detail two user studies. The corresponding image results are shown in Figures \ref{fig:experience1} and \ref{fig:experience2}, while the task results from the user studies are presented in Figures \ref{fig:sup_user_study1} and \ref{fig:sup_user_study2}. A total of 98 participants contributed 490 votes.

For the Image-to-3D task, we referred to Figure \ref{fig:experience1} to present comparison images to the participants and asked them to answer the following question: "Please select the best 3D model reconstructed from an input 2D image, generated by using ATIH to fuse the front-view image of the given 3D object model with another object text." For the 3D-to-3D task, we referred to Figure \ref{fig:experience2} to present comparison images, and participants were instructed to answer: "Please select the best 3D model generated by fusing a 3D object model with another object text. When making your selection, focus on the model's 3D fusion and artistic expressiveness."

\begin{table*}[htbp]
\vskip -0in
\centering
\setlength{\tabcolsep}{7pt}
\renewcommand{\arraystretch}{1.1}
\caption{User study with Image-to-3D methods.}
\resizebox{0.90\linewidth}{!}{
\begin{tabular}{c||c|c|c|c|c}
\Xhline{1.2pt} 
\diagbox{3D model-prompt}{options(Models)} & A(Our C33D) & B(LGM) & C(Vfusion3D) & D(Era3D) & E(CRM)\\ 
\hline
cartoon cat-dock  & 71.43\% & 4.08\% & 6.12\% & 17.35\% & 1.02\% \\
cartoon lion-egyptian cat  & 89.80\% & 3.06\% & 2.04\% & 4.08\% & 1.02\% \\
pole finial-fern  & 46.94\% & 11.22\% & 0.00\% & 35.71\% & 6.12\% \\
teddy bear-greenhouse  & 74.49\% & 6.12\% & 1.02\% & 8.16\% & 10.20\% \\
goblet-silver  & 79.59\% & 0.00\% & 0.00\% & 2.04\% & 18.37\% \\
\Xhline{1.2pt}
\end{tabular}}

\label{tab:sup_votes1}
\end{table*}

\begin{table*}[htbp]
\vskip -0in
\centering
\setlength{\tabcolsep}{7pt}
\renewcommand{\arraystretch}{1.1}
\caption{User study with 3D-to-3D method.}
\resizebox{0.9\linewidth}{!}{
\begin{tabular}{c||c|c|c|c}
\Xhline{1.2pt} 
\diagbox{3D model-prompt}{options(Models)} & A(Our C33D)& B(Themestation)  & C(Phidias) & D(Mvedit) \\ 
\hline
gargoyle statue-gecko  & 69.39\% & 5.10\% & 20.41\% & 5.10\% \\
cartoon crow-flamingo  & 80.61\% & 13.27\% & 3.06\% & 3.06\% \\
cartoon cow-grape & 67.35\% & 14.29\% & 16.33\% & 2.04\% \\
cartoon cannon-aluminum  & 56.12\% & 1.02\% & 38.78\% & 4.08\% \\
jaguar vessel-Screwdriver  & 94.90\% & 0.00\% & 3.06\% & 2.04\% \\
\Xhline{1.2pt}
\end{tabular}}

\label{tab:sup_votes2}
\end{table*}

\section{More Ablation Study}
\label{sec:More Ablaution}

We conduct additional ablation studies to evaluate the impact of three core components in our pipeline: (1) initialization with multi-view 3D features, (2) integration of the front-view image with time embedding, and (3) our Fusion-guided Adaptive Inversion (FAI). 

\begin{figure}[t]
  \centering
\includegraphics[width=1.0\linewidth]{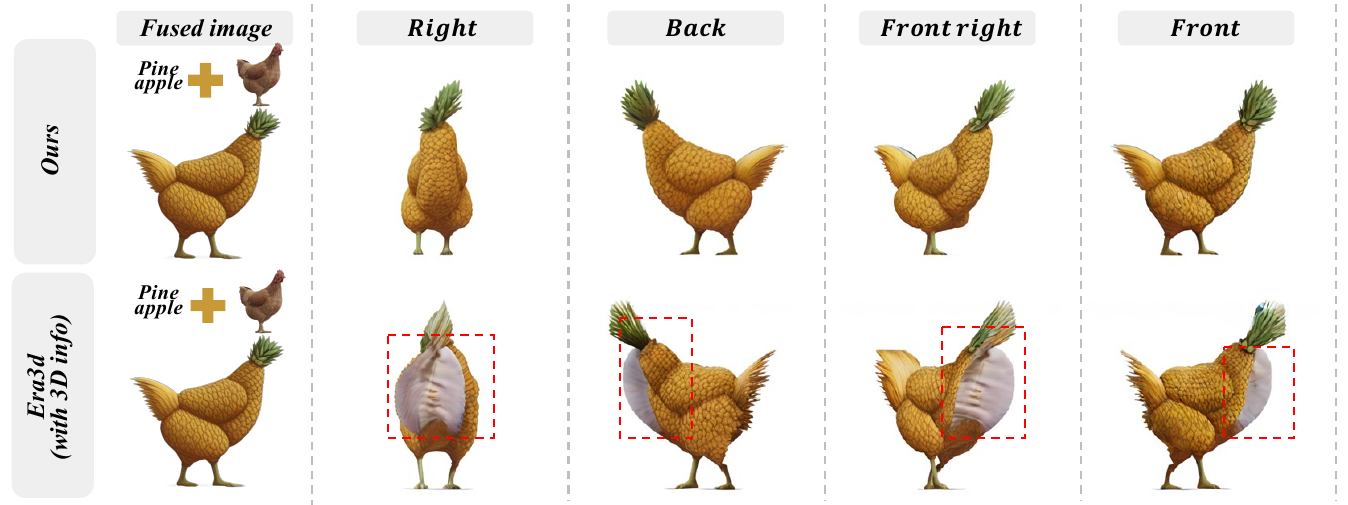}
 \vskip -0in
\caption{Qualitative comparison in the ablation study. Ours (top) vs. Era3D with 3D features (bottom). Red boxes indicate distortions and inconsistencies. }
    \label{fig:era3d_with_3Dinfo}
\end{figure}

\textbf{Multi-view feature initialization.}  
To test whether directly using 3D-aware features benefits the generation process, we reinitialize \textbf{Era3D} with multi-view features extracted from the original model. These features are noised to a specific timestep and denoised using the standard diffusion process. As shown in Fig.~\ref{fig:era3d_with_3Dinfo}, this leads to visible geometric distortions and inconsistent textures across different views. In contrast, our method yields more stable shapes and coherent appearance, indicating that our inversion initialization strategy provides more effective conditioning than reusing raw 3D features.

\begin{figure}[t]
  \centering
\includegraphics[width=1.0\linewidth]{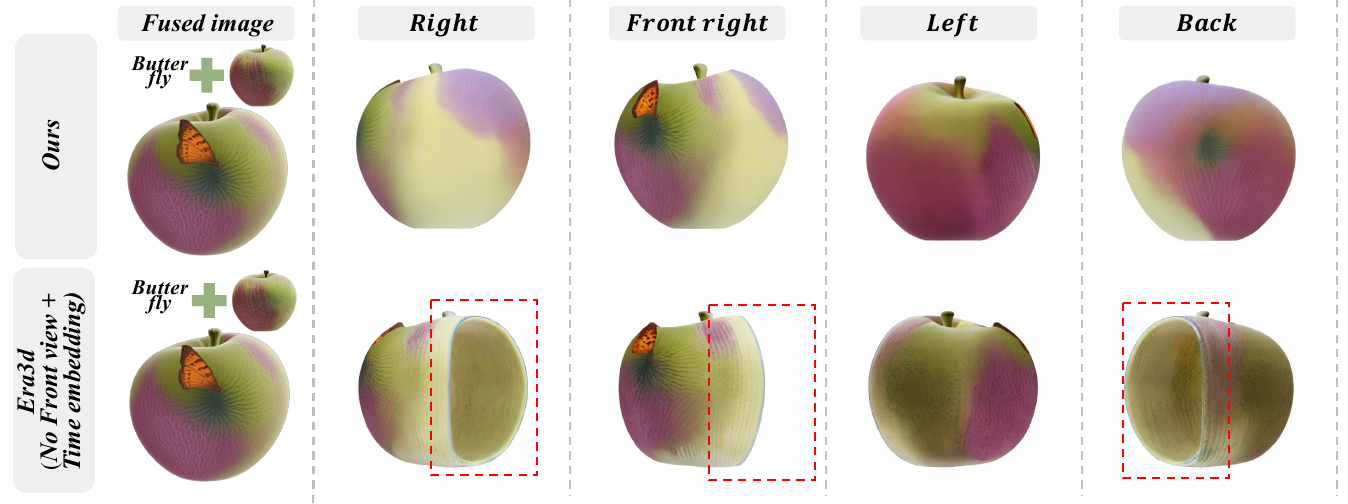}
 \vskip -0in
\caption{Ablation study on the integration of front-view image and time embedding. Top: our full model. Bottom: without front-view + time embedding. Red boxes highlight inconsistencies and blurred regions. }
    \label{fig:No_Front_view_Time_embedding}
\end{figure}

\textbf{Front-view and time embedding.}  
We investigate the importance of combining the front-view image with time embedding during the fusion process. When this component is removed, the model loses consistent guidance during denoising, resulting in significant artifacts and blurry textures—particularly in non-frontal views—as shown in Fig.~\ref{fig:No_Front_view_Time_embedding}. This highlights the necessity of front-view temporal conditioning to maintain geometric alignment and texture consistency across multiple views.

\textbf{Forward noise vs. FAI.}  
We also compare our Fusion-guided Adaptive Inversion (FAI) with a simpler baseline that applies random forward noising to the clean latent, without structure-aware control. As depicted in Fig.~\ref{fig:noise_add_clean_compare}, this baseline suffers from over-smoothing. In contrast, our FAI approach allows controlled information retention and produces higher-fidelity, view-consistent results, validating the advantage of adaptive, fusion-aware inversion in our pipeline.

\begin{figure}[t]
  \centering
\includegraphics[width=1.0\linewidth]{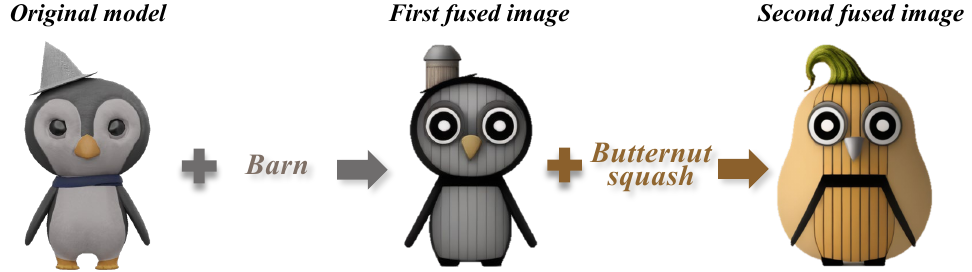}
 \vskip -0in
\caption{Progressive multi-category fusion: the original 3D model is first fused with ``barn'' and then further enriched with ``Butternut squash'' characteristics, showcasing our framework's compositional capability. }
    \label{fig:double_fuse}
\end{figure}

\begin{figure}[t]
  \centering
\includegraphics[width=1.0\linewidth]{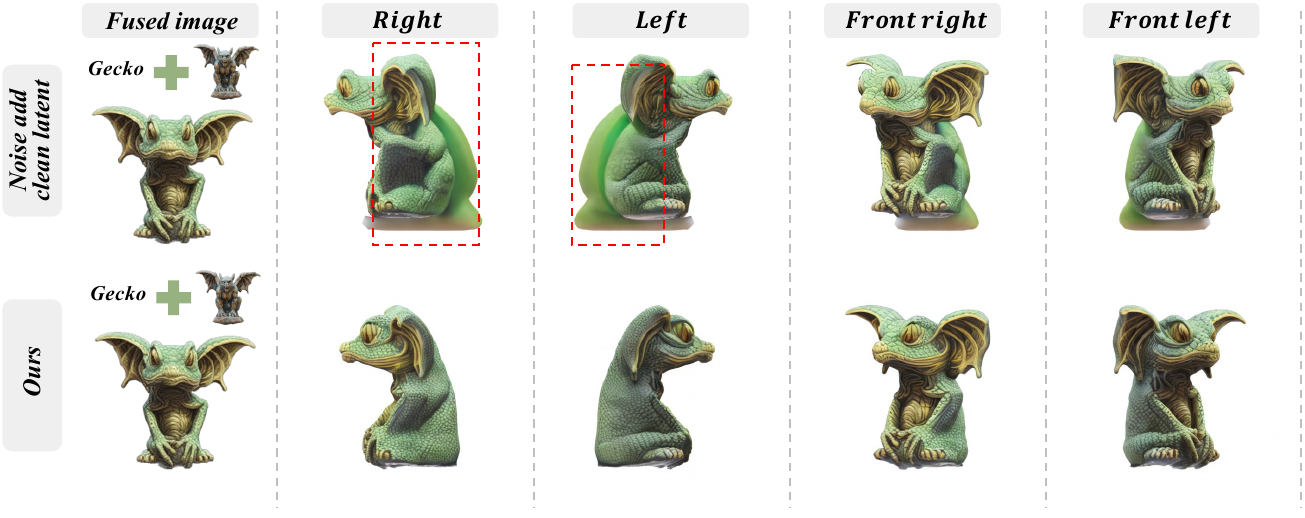}
 \vskip -0in
\caption{Ablation study comparing our Fusion-guided Adaptive Inversion (FAI) to simple forward noising. The baseline (top) suffers from shape distortion and texture inconsistency. Our method (bottom) better preserves geometry and cross-view coherence.}
    \label{fig:noise_add_clean_compare}
\end{figure}

\section{More Comparisons}
\label{sec:More Comparisons}

\begin{figure*}[t]
  \centering
 \includegraphics[width=0.95\linewidth]{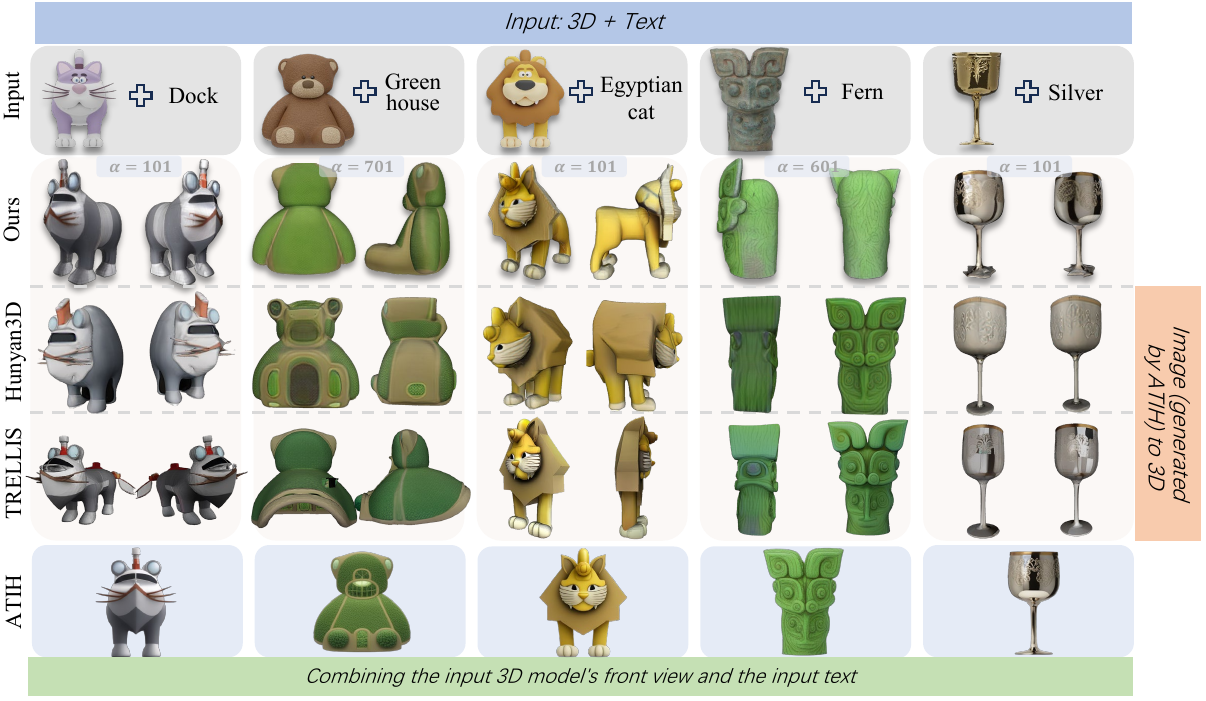}
   \caption{\textbf{Comparisons with 3D-to-3D methods.}  We observe that our method better preserves object structure and fusion semantics, while Trellis and Hunyuan3D suffer from geometry collapse and inconsistent features.}
   \label{fig:more_experience_image23d}
\end{figure*}

We conducted comparisons with Trellis \cite{xiang2025structured} and Hunyuan3D \cite{yang2024hunyuan3d}. As illustrated in Fig.~\ref{fig:more_experience_image23d}, both baseline methods exhibit common failure patterns, including \textit{Janus artifacts}, \textit{geometric distortions}, and \textit{inconsistent feature fusion}, particularly under challenging compositional scenarios (e.g., ``teddy bear + greenhouse''). These observations are consistent with the limitations previously identified in Era3D (Fig.~6), further highlighting the robustness of our approach in preserving structural integrity and semantic consistency across views.

\section{More results}
\label{sec:More results}

\begin{figure*}[h]
  \centering
 \includegraphics[width=0.9\linewidth]{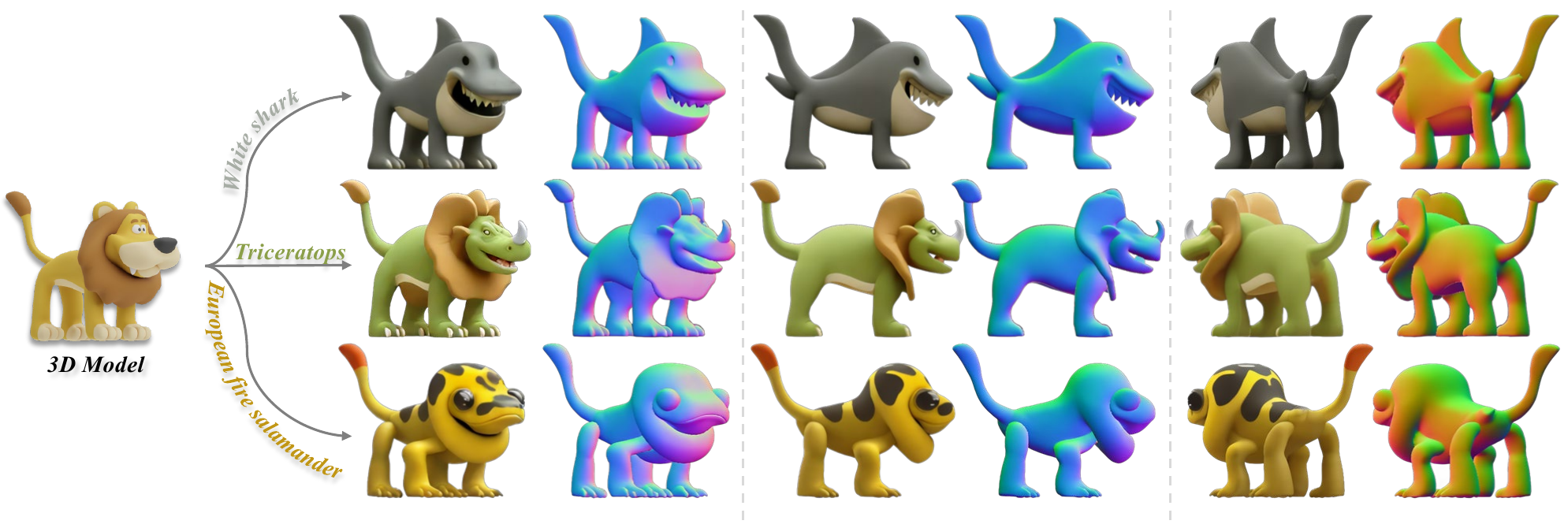}
   \caption{more results}
   \label{fig:more_results_lion}
\end{figure*}

\begin{figure*}[h]
  \centering
 \includegraphics[width=0.9\linewidth]{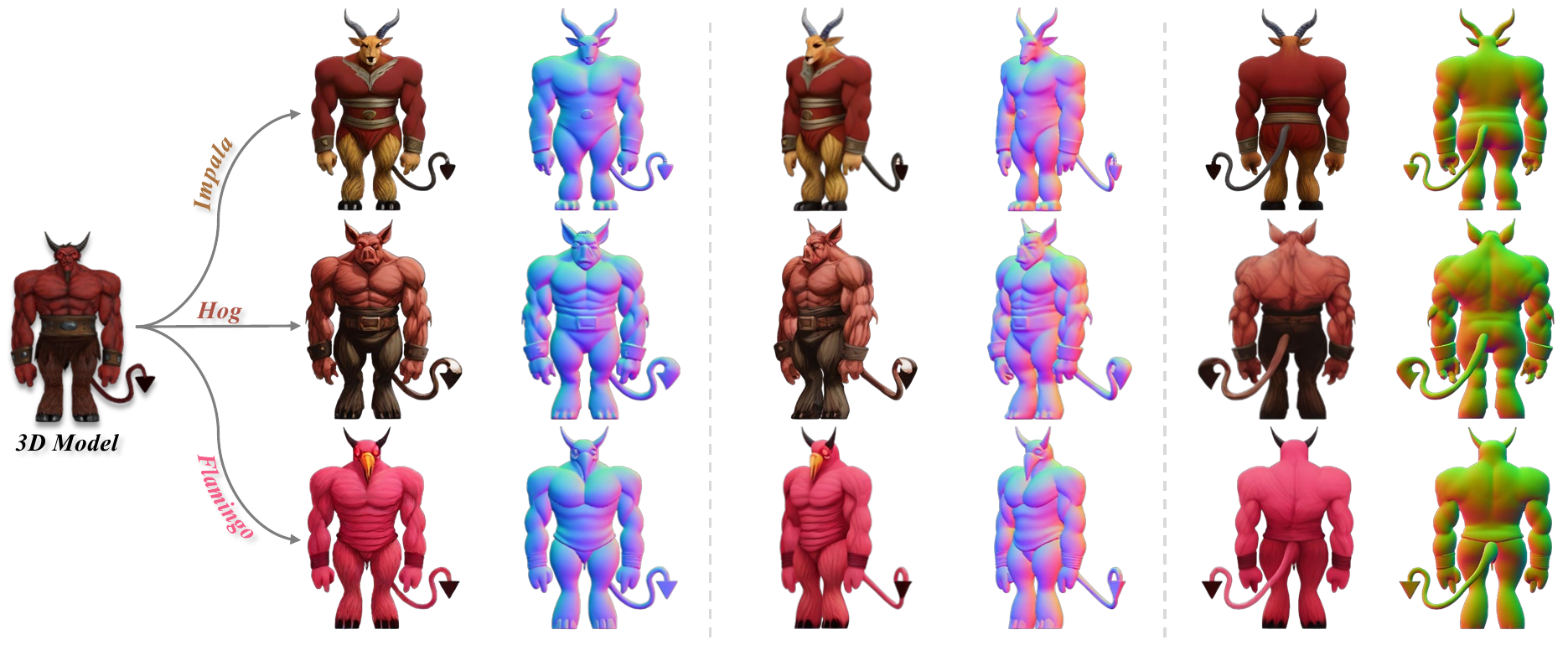}
   \caption{more results}
   \label{fig:more_results_devill}
\end{figure*}

\begin{figure*}[h]
  \centering
 \includegraphics[width=0.9\linewidth]{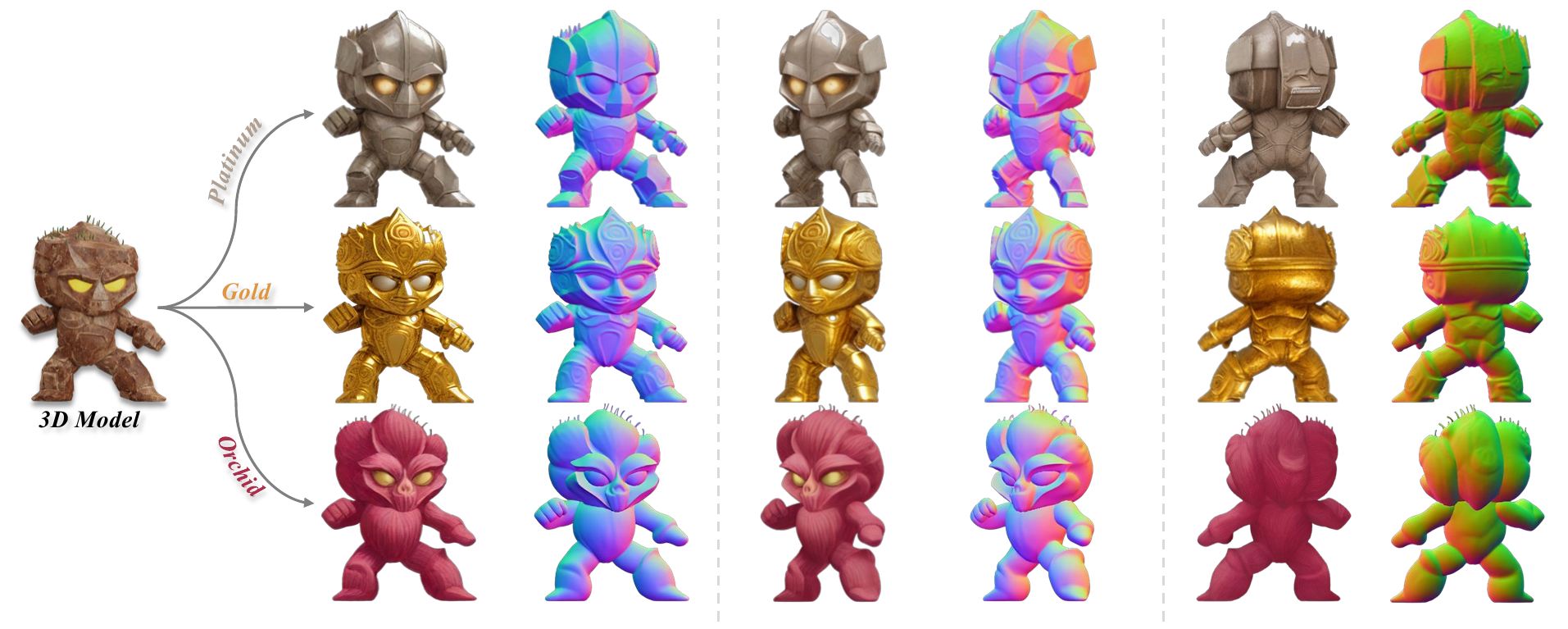}
   \caption{more results}
   \label{fig:more_results_kinn}
\end{figure*}

\begin{figure*}[h]
  \centering
 \includegraphics[width=0.9\linewidth]{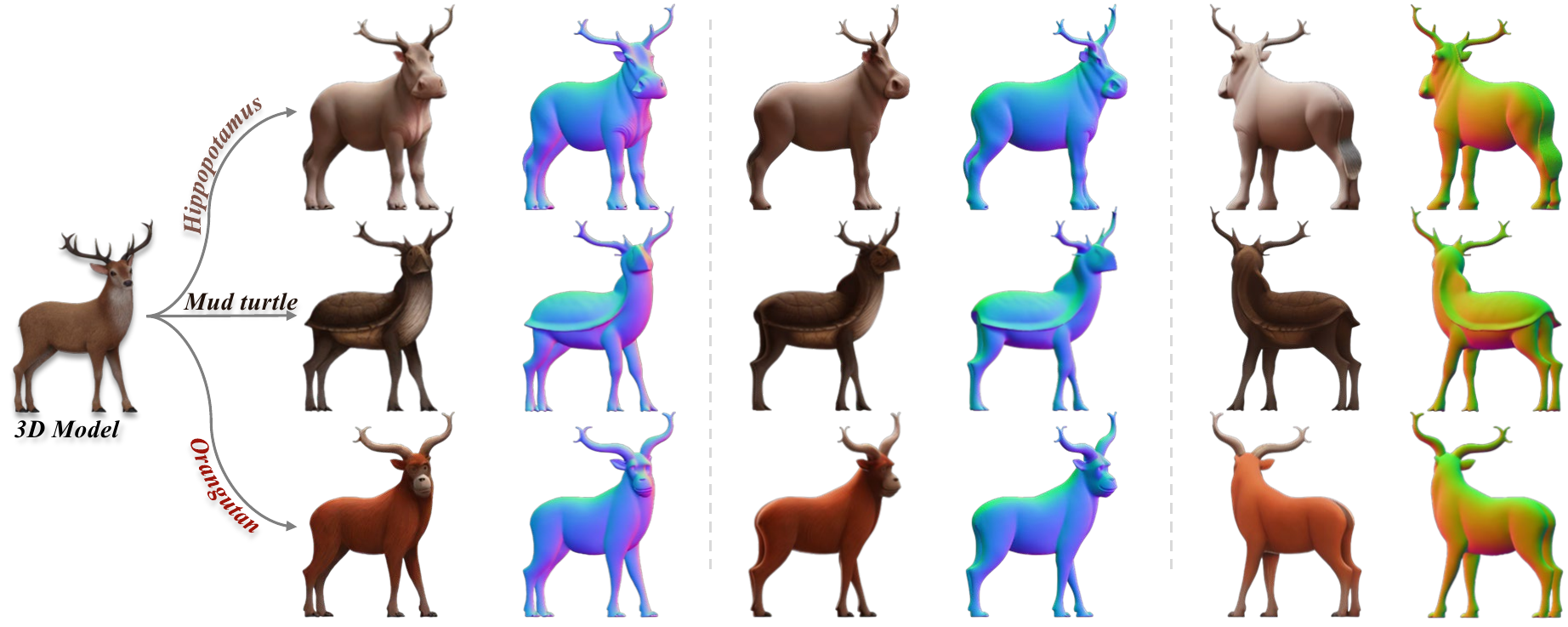}
   \caption{more results}
   \label{fig:more_results_deer}
\end{figure*}

\begin{figure*}[h]
  \centering
 \includegraphics[width=0.9\linewidth]{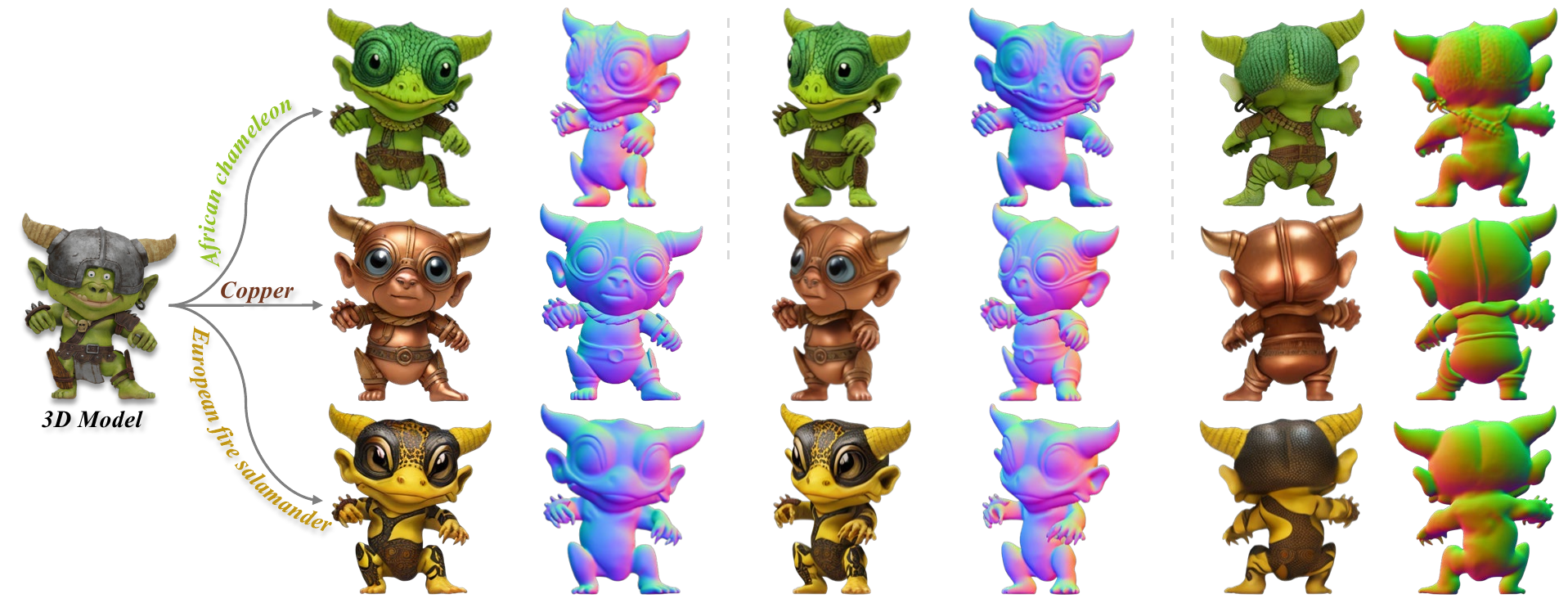}
   \caption{more results}
   \label{fig:more_results_smoll}
\end{figure*}

\begin{figure*}[h]
  \centering
 \includegraphics[width=0.9\linewidth]{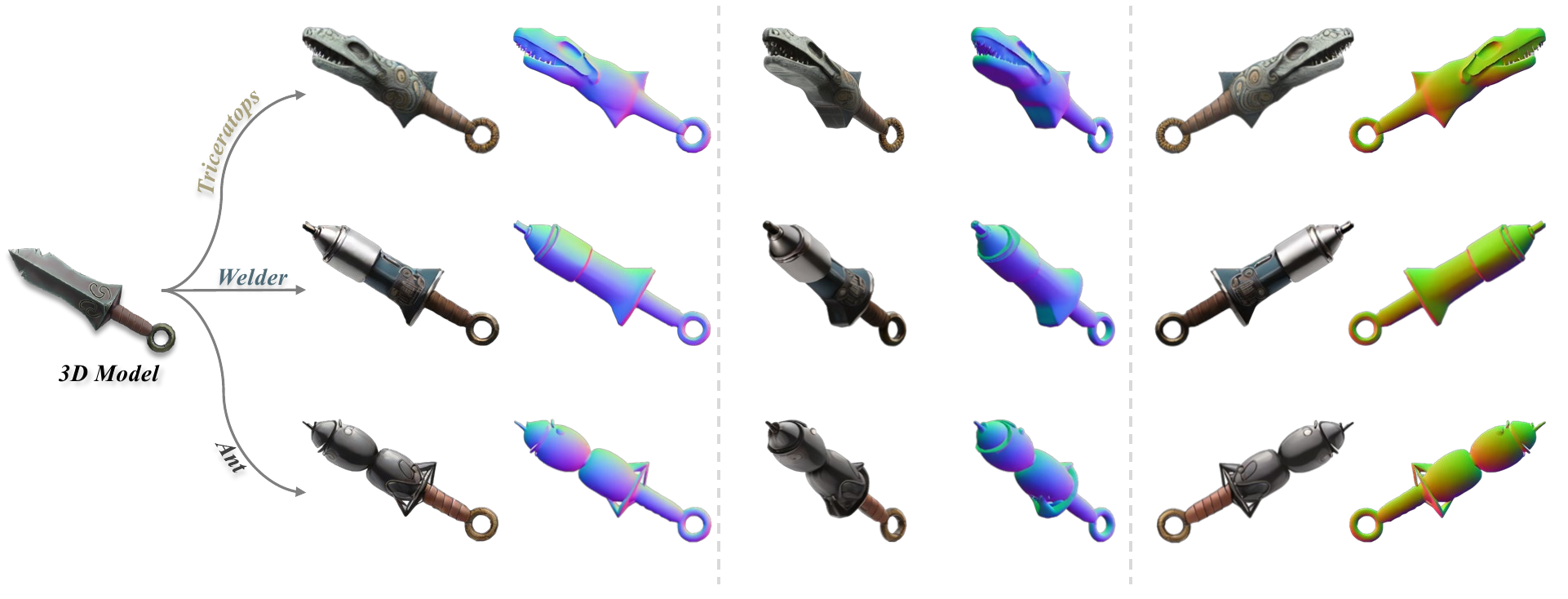}
   \caption{more results}
   \label{fig:more_results_sword}
\end{figure*}

\begin{figure*}[htbp]
  \centering
\includegraphics[width=0.42\linewidth]{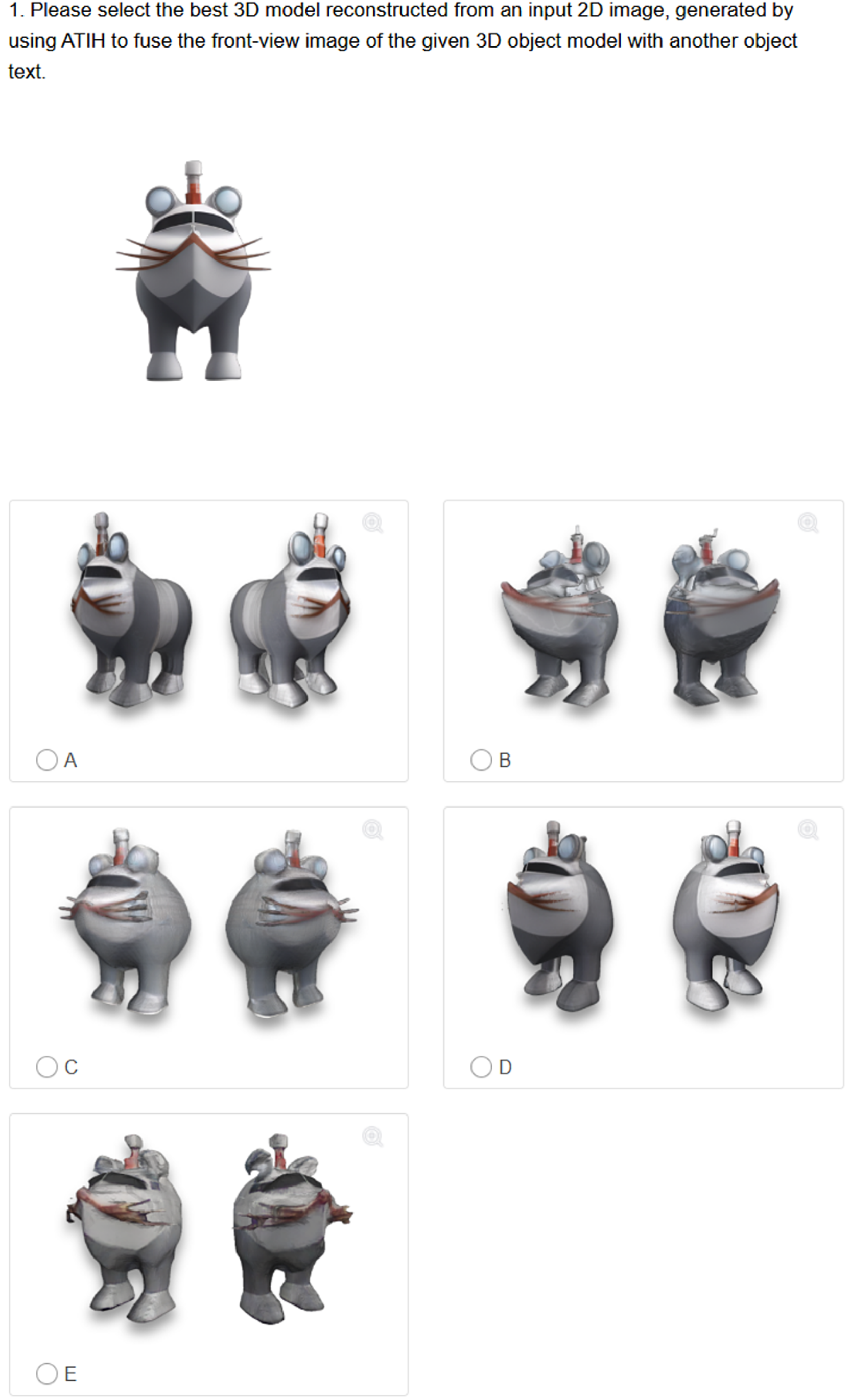}
\caption{An example of a user study comparing various Image-to-3D methods. }
    \label{fig:sup_user_study1}
\end{figure*}

\begin{figure*}[htbp]
  \centering
\includegraphics[width=0.4\linewidth]{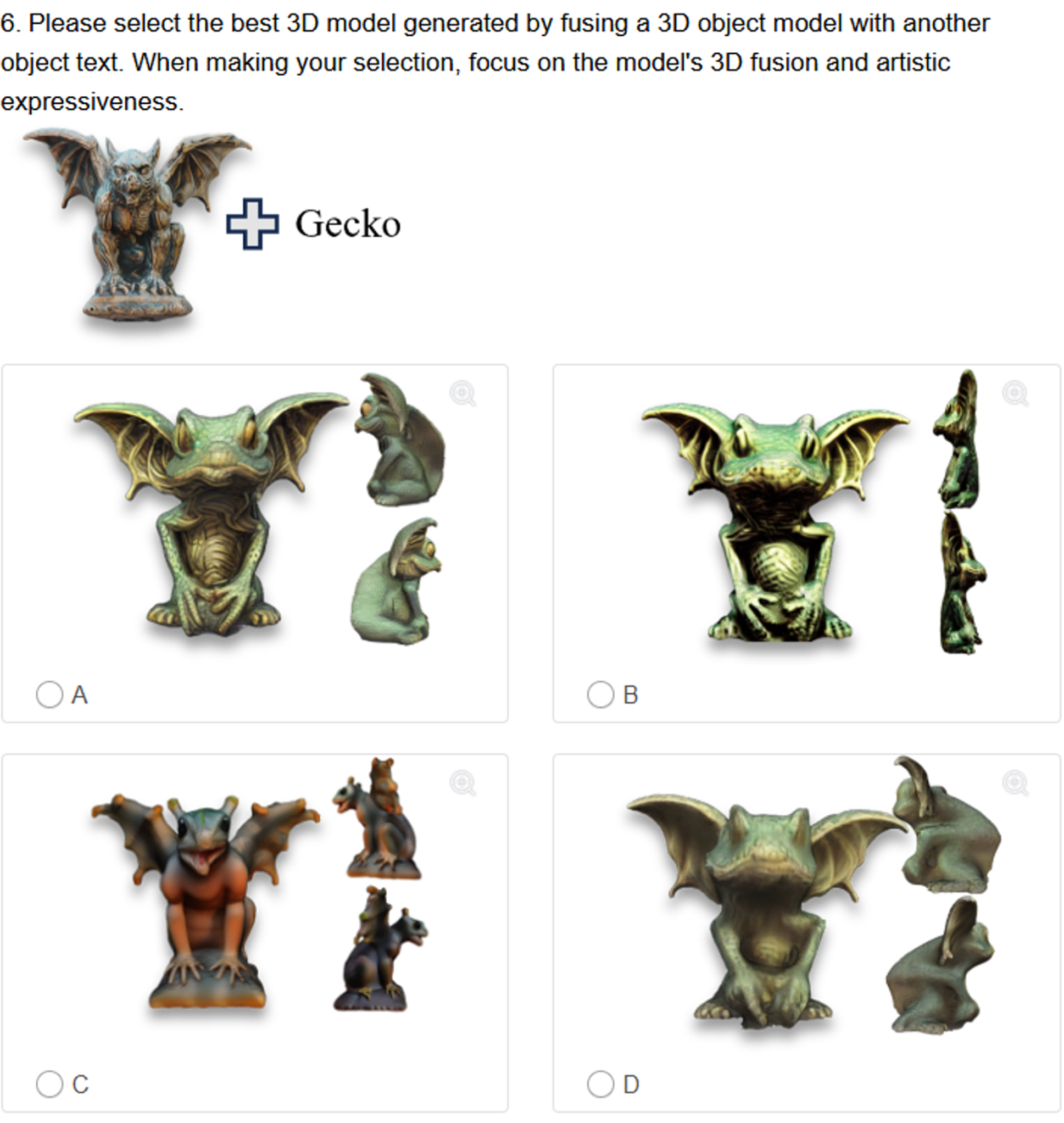}
\caption{An example of a user study comparing  3D-to-3D method. }
    \label{fig:sup_user_study2}
\end{figure*}

In this section, we present additional experimental results to further demonstrate the effectiveness and versatility of our C33D framework. Figure~\ref{fig:more_results_lion}, Figure~\ref{fig:more_results_deer}, Figure~\ref{fig:more_results_kinn}, Figure~\ref{fig:more_results_smoll}, Figure~\ref{fig:more_results_devill}, and Figure~\ref{fig:more_results_sword} showcase multi-view outputs generated from various types of 3D models fused with diverse text prompts. These results illustrate our model’s ability to integrate textual semantics with 3D geometry in a structurally coherent and visually creative manner.

Furthermore, our framework is inherently extensible to more complex fusion scenarios involving multiple target concepts (e.g., ``3D model + Category A + Category B''). By modifying the initial 2D fusion step (ATIH) to accommodate multiple textual descriptions, our method enables progressive and compositional synthesis. As shown in Figure~\ref{fig:double_fuse}, we demonstrate this capability by fusing a cartoon penguin with both ``Barn'' and ``Butternut squash'' features, resulting in a harmonious blend that retains key characteristics from each category. This further highlights the generalization and compositional flexibility of our approach in generating novel 3D content.


\end{document}